\documentclass[10pt,twocolumn,letterpaper]{article}

\usepackage[pagenumbers]{cvpr} % To force page numbers, e.g. for an arXiv version

\usepackage{graphicx}
\usepackage{amsmath}
\usepackage{amssymb}
\usepackage{booktabs}
\usepackage{adjustbox}
\usepackage{multirow}
\usepackage{graphicx}
\usepackage{pifont}
\usepackage{gensymb}
\usepackage{float}
\usepackage[usenames,dvipsnames]{color}
\usepackage{colortbl} % For cell coloring

\usepackage{background}

\newcommand{\mywatermark}{%
    \begin{minipage}{\textwidth}
        \centering
        \fontsize{10}{10}\selectfont % Adjust font size and baselineskip
        This paper has been accepted for publication at the IEEE Conference on Computer Vision and Pattern Recognition (CVPR), Seattle, 2024. ©IEEE
    \end{minipage}%
}

\backgroundsetup{
    scale=1,
    color=gray,
    angle=0,
    opacity=0.5,
    position=current page.north,
    vshift=-1cm, % Adjust vertical position
    contents={\mywatermark}
}

\definecolor{highlight}{rgb}{0.9, 0.9, 0.9} % Light gray
\definecolor{highlight2}{rgb}{0.9, 0.9, 0.95} % Light gray
\definecolor{highlight3}{rgb}{0.95, 0.9, 0.9} % Light gray

\usepackage[accsupp]{axessibility}

\usepackage{soul}
\usepackage{lipsum}  

\usepackage[pagebackref,breaklinks,colorlinks]{hyperref}

\usepackage[capitalize]{cleveref}
\crefname{section}{Sec.}{Secs.}
\Crefname{section}{Section}{Sections}
\Crefname{table}{Table}{Tables}
\crefname{table}{Tab.}{Tabs.}

\def\cvprPaperID{16345} 
\def\confName{CVPR}
\def\confYear{2024}

\usepackage{fancyhdr}

\begin{document}

\title{\textit{eTraM}: Event-based Traffic Monitoring Dataset}

\author{Aayush Atul Verma$^*$, Bharatesh Chakravarthi$^*$, Arpitsinh Vaghela$^*$, Hua Wei, Yezhou Yang \\
Arizona State University \\
{\tt\small \{averma90, bshettah, avaghel3, hua.wei, yz.yang\}@asu.edu}\
}
\twocolumn[{%
\renewcommand\twocolumn[1][]{#1}%
\maketitle

\begin{center}
\captionsetup{type=figure}\includegraphics[width=0.95\textwidth]{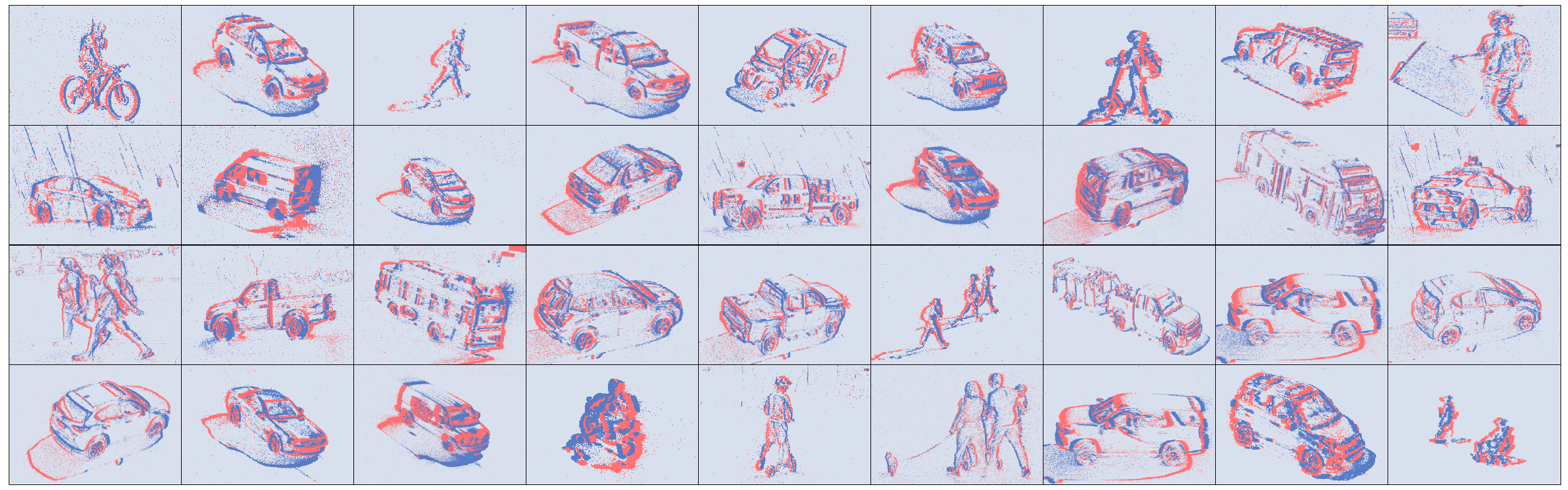}
\captionof{figure}{Unveiling the Dynamic World of Road Traffic: A glimpse into our event-based traffic monitoring dataset featuring diverse traffic participants including pedestrians, various sized vehicles, and micro-mobility that include cycles, wheelchairs, and e-scooters.}
\label{fig01}
\end{center}%
}]

\def\thefootnote{*}\footnotetext{Equal contribution}
%%%%%%%%% ABSTRACT
\cfoot{Footer}

\begin{abstract} Event cameras, with their high temporal and dynamic range and minimal memory usage, have found applications in various fields. However, their potential in static traffic monitoring remains largely unexplored. To facilitate this exploration, we present eTraM - a first-of-its-kind, fully event-based traffic monitoring dataset. eTraM offers $10\,\text{hr}$ of data from different traffic scenarios in various lighting and weather conditions, providing a comprehensive overview of real-world situations. Providing 2M bounding box annotations, it covers eight distinct classes of traffic participants, ranging from vehicles to pedestrians and micro-mobility. eTraM's utility has been assessed using state-of-the-art methods for traffic participant detection, including RVT, RED, and YOLOv8. We quantitatively evaluate the ability of event-based models to generalize on nighttime and unseen scenes. Our findings substantiate the compelling potential of leveraging event cameras for traffic monitoring, opening new avenues for research and application. eTraM is available at \url{https://eventbasedvision.github.io/eTraM}.
\end{abstract}

%%%%%%%%% BODY TEXT
\section{Introduction}
\label{sec:intro}

In the dynamic landscape of modern transportation, Intelligent Transportation Systems (ITS) play a pivotal role in enhancing traffic flow~\cite{zhou2020enhancing, mei2023libsignal}, route optimization~\cite{liu2013adaptive, wei2021we}, and safety~\cite{garcia2021safety, du2023safelight}. As an essential task in ITS, traffic participant detection aims to provide information assisting in counting, measuring speed, identifying traffic incidents, and predicting traffic flow. The detection methods must be fast enough to operate in real-time and be insensitive to illumination change and varying weather conditions while utilizing less storage. For instance, in the span of just $1\,\text{s}$, a vehicle may travel over $8.3\,\text{m}$, and a pedestrian could cover over $1.43\,\text{m}$, leading to potential misses in fast-paced traffic scenarios and introducing motion blur concerns~\cite{zhang2023vehicle}. Moreover, nighttime and different weather conditions make the detection task more challenging as many features such as edge, corner, and shadow do not work due to varying illumination. In the face of these challenges, the integration of event cameras into ITS holds great promise for robust traffic participant detection in real-time scenarios.  

Event cameras capture an asynchronous and continuous stream of ``events" or pixel-level brightness changes instead of traditional static frames at fixed frequencies. Each event is represented by a tuple $\langle x, y, p, t \rangle$ corresponding to an illuminance change greater than a fixed threshold at pixel location $(x, y)$ and time $t$, with the polarity $p\in\{+1, -1\}$ indicating whether the illuminance increased or decreased. Event cameras with their exceptional temporal resolution (over $10,000\,\text{fps}$) and high dynamic range (above $120\,\text{dB}$)~\cite{8946715} have prompted explorations into visual perception, robotics~\cite{eventvisionsurvey, eventvisionsurvey2}, and its various applications in ITS~\cite{UASMonitoring, 9812059, eventvisionsurvey}.

Existing multimodal traffic datasets captured from various sensors, including RGB cameras, LiDAR, and Radar, have been utilized for several tasks in the context of autonomous vehicles (AV)~\cite{kitti, waymo, nuscenes, AVevent}. However, a largely unexplored yet promising domain lies in the use of event cameras for detection in traffic monitoring. This serves as an inspiration for us to contribute a first-of-its-kind, fully event-based traffic monitoring dataset. 

In this paper, we present \textit{eTraM} a novel fully event-based traffic perception dataset curated using the state-of-the-art (SOTA) high-resolution Prophesee EVK4 HD event camera~\cite{prophesee-evk4}. The dataset spans over $10\,\text{hr}$ of annotated data from a static perspective that facilitates comprehensive traffic monitoring. We consulted experts from the Institute of Automated Mobility (IAM)~\cite{iam} and 
strategically mounted an event camera over selected sites (intersections, roadways, and local streets) to collect traffic data under diverse conditions. The data collection process was systematically conducted across various weather and lighting conditions spanning challenging scenarios such as high glare, overexposure, underexposure, nighttime, twilight, and rainy days. \textit{eTraM} includes over $2\text{M}$  bounding box annotations of traffic participants such as vehicles (cars, trucks, buses, trams), pedestrians, and various micro-mobility (bikes, bicycles, wheelchairs) as shown in \figureautorefname~\ref{fig01}. 

\textit{eTraM} offers the perspective of a static event camera captured at various scenes, further enhancing its versatility and applicability in real-world scenarios. This approach ensures that \textit{eTraM} captures not only the routine dynamics of traffic but the nuances and challenges presented by a broad spectrum of scenarios and participants as well. We tested the diversity and quality of the dataset through various experiments and evaluated the generalization of event-based methods on nighttime and unseen scenes. \textit{eTraM} stands as a valuable resource, propelling research and innovation in the evolving field of event-based traffic perception in ITS. Our contributions can be summarized as follows:
\begin{enumerate}
\item A first-of-its-kind fully event-based dataset from a static perspective encompassing a diverse variety of traffic scenarios (scenes, weather, and lighting conditions) and participants (vehicles, pedestrians, and micro-mobility).
\item Establish baselines using state-of-the-art event-based methods on \textit{eTraM} across various traffic monitoring scenes and lighting conditions.
\item Quantitative evaluation of the generalization capabilities of event-based methods on nighttime and unseen scenarios.
\end{enumerate}

The rest of this paper is structured as follows. Section~\ref{sec:related-datasets} briefly summarizes the existing event-based dataset contributions. Section~\ref{sec:dataset} presents \textit{eTraM} dataset with details on the acquisition framework, preprocessing, and annotation. In Section~\ref{sec:eval}, we provide baseline results and explore the generalization capabilities of event-based methods on nighttime and unseen scenarios. Section~\ref{sec:discussion} highlights the advantages of utilizing event cameras and the difference between static and ego-motion event datasets. 

%-------------------------------------------------------------------------
\section{Related Datasets}
\label{sec:related-datasets}

\begin{table*}[t!]
\centering
\resizebox{0.97\linewidth}{!}{%
\normalfont
\begin{tabular}
{l|c|c|cc|ccc|ccc|cc|r|l}
\toprule \midrule
\multicolumn{1}{c|}{\multirow{2}{*}{\textbf{\begin{tabular}[c]{@{}c@{}} Dataset \\ Name\end{tabular}}}} &
  \multirow{2}{*}{\textbf{Year}} &
\multirow{2}{*}{\textbf{\begin{tabular}[c]{@{}c@{}}Duration\\ (in hours) \end{tabular}}} &  \multicolumn{2}{c|}{\textbf{Perspective}} &
  \multicolumn{3}{c|}{\textbf{\begin{tabular}[c]{@{}c@{}}Traffic \\ Participants\end{tabular}}} &
  \multicolumn{3}{c|}{\textbf{Lighting}} &
  \multicolumn{2}{c|}{\textbf{Weather}} &
  \multirow{2}{*}{\textbf{\begin{tabular}[c]{@{}c@{}}No. of\\ Bbox \end{tabular}}} &
  \multicolumn{1}{c}{\multirow{2}{*}{\textbf{Scenarios}}} \\  \cline{4-13}
\multicolumn{1}{c|}{} &
   &
   &
  \multicolumn{1}{c|}{\textbf{Ego}} &
  \textbf{Static} &
  \multicolumn{1}{c|}{\textbf{VH}} &
  \multicolumn{1}{c|}{\textbf{PED}} &
  \textbf{MM} &
  \multicolumn{1}{c|}{\textbf{Day}} &
  \multicolumn{1}{c|}{\textbf{Night}} &
  \textbf{Twilight} &
  \multicolumn{1}{c|}{\textbf{Clear}} &
  \textbf{Rainy} &
   &
  \multicolumn{1}{c}{} \\ \midrule \midrule
\textbf{DDD17 \cite{ddd17}} &
  2017 &
  12 &
  \multicolumn{1}{c|}{\checkmark} & 
   & 
  \multicolumn{1}{c|}{\checkmark} &
  \multicolumn{1}{c|}{} &
   &
  \multicolumn{1}{c|}{\checkmark} &
  \multicolumn{1}{c|}{\checkmark} &
  \checkmark &
  \multicolumn{1}{c|}{\checkmark} &
  \checkmark &
  \multicolumn{1}{c|}{-} &
  Driving \\
  \textbf{MVSEC \cite{mvsec}} &
  2018 &
   \multicolumn{1}{c|}{-} &
  \multicolumn{1}{c|}{} &
   &
  \multicolumn{1}{c|}{\checkmark} &
  \multicolumn{1}{c|}{} &
   &
  \multicolumn{1}{c|}{\checkmark} &
  \multicolumn{1}{c|}{\checkmark} &
   &
  \multicolumn{1}{c|}{\checkmark} &
   &
  \multicolumn{1}{c|}{-} &
  Driving, Handheld \\
\textbf{DVS Pedestrian} \cite{dvspedestrian}&
  2019 &
  0.1 &
  \multicolumn{1}{c|}{} &
  \checkmark &
  \multicolumn{1}{c|}{} &
  \multicolumn{1}{c|}{\checkmark} &
   &
  \multicolumn{1}{c|}{\checkmark} &
  \multicolumn{1}{c|}{} &
   &
  \multicolumn{1}{c|}{\checkmark} &
  \checkmark &
  4.6K &
  Walking street \\
\textbf{DDD20 \cite{ddd20}} &
  2020 &
  51 &
  \multicolumn{1}{c|}{\checkmark} &
   &
  \multicolumn{1}{c|}{\checkmark} &
  \multicolumn{1}{c|}{} &
   &
  \multicolumn{1}{c|}{\checkmark} &
  \multicolumn{1}{c|}{\checkmark} &
  \checkmark &
  \multicolumn{1}{c|}{\checkmark} &
   &
  \multicolumn{1}{c|}{-} &
  Driving \\
\textbf{Gen1} \cite{gen1}&
  2020 &
  39 &
  \multicolumn{1}{c|}{\checkmark} &
   &
  \multicolumn{1}{c|}{\checkmark} &
  \multicolumn{1}{c|}{\checkmark} &
   &
  \multicolumn{1}{c|}{\checkmark} &
  \multicolumn{1}{c|}{} &
   &
  \multicolumn{1}{c|}{\checkmark} &
   &
  255K &
  Driving \\
\textbf{1 Megapixel \cite{1megapixel}} &
  2020 &
  15 &
  \multicolumn{1}{c|}{\checkmark} &
   &
  \multicolumn{1}{c|}{\checkmark} &
  \multicolumn{1}{c|}{\checkmark} &
  \checkmark &
  \multicolumn{1}{c|}{\checkmark} &
  \multicolumn{1}{c|}{\checkmark} &
   &
  \multicolumn{1}{c|}{\checkmark} &
   &
  25M &
  Driving \\
\textbf{DSEC \cite{dsec}} &
  2021 &
  1 &
  \multicolumn{1}{c|}{\checkmark} &
   &
  \multicolumn{1}{c|}{\checkmark} &
  \multicolumn{1}{c|}{\checkmark} &
  \checkmark &
  \multicolumn{1}{c|}{\checkmark} &
  \multicolumn{1}{c|}{\checkmark} &
   &
  \multicolumn{1}{c|}{\checkmark} &
   &
  390K &
  Driving \\
  \textbf{DVS-OUTLAB \cite{dvsoutlab}} &
  2021 &
  7 &
  \multicolumn{1}{c|}{} &
  \checkmark &
  \multicolumn{1}{c|}{} &
  \multicolumn{1}{c|}{} &
  \checkmark &
  \multicolumn{1}{c|}{\checkmark} &
  \multicolumn{1}{c|}{} &
   &
  \multicolumn{1}{c|}{\checkmark} &
  \checkmark &
  $\sim$ 47K &
  Playground \\

\textbf{PEDRo \cite{pedro}} &
  2023 &
  0.5 &
  \multicolumn{1}{c|}{\checkmark} &
   &
  \multicolumn{1}{c|}{} &
  \multicolumn{1}{c|}{\checkmark} &
   &
  \multicolumn{1}{c|}{\checkmark} &
  \multicolumn{1}{c|}{\checkmark} &
  \checkmark &
  \multicolumn{1}{c|}{\checkmark} &
  \checkmark &
  43K &
  Robotics \\ \midrule \midrule
\textbf{\textit{eTraM} (Ours)} &
  2024 &
  10 &
  \multicolumn{1}{c|}{\checkmark} &
  \checkmark &
  \multicolumn{1}{c|}{\checkmark} &
  \multicolumn{1}{c|}{\checkmark} &
  \checkmark &
  \multicolumn{1}{c|}{\checkmark} &
  \multicolumn{1}{c|}{\checkmark} &
  \checkmark &
  \multicolumn{1}{c|}{\checkmark} &
  \checkmark &
  $\sim$ 2M
   &
  \begin{tabular}[c]{@{}l@{}}Intersections, Roadways,\\ Local streets\end{tabular} \\ \midrule \bottomrule
    \end{tabular}
}
\caption{A comprehensive overview of event-based traffic datasets from 2017 to 2024. (VH - Vehicle, PED - Pedestrian, MM - Micro-mobility) 
}
\label{tab:related_datasets}
\end{table*}

\textbf{Early event-based datasets} often involved the transformation of frame-based datasets into event streams. A noteworthy example is~\cite{orchard2015converting}, where MNIST~\cite{726791} and Caltech-101~\cite{1597116} datasets were converted to event streams by moving an event camera in front of a screen displaying frames. Later works utilized event simulators~\cite{esim, v2e} to convert widely-used frame-based datasets into their event-based counterparts. 

\textbf{Ego-motion event-based datasets} have seen a rise in recent years due to the increased accessibility of event sensors. DDD17~\cite{ddd17}, DDD20~\cite{ddd20} pioneered the initial efforts in deploying event cameras for driving scenarios using a $346\times260\,\text{px}$ DAVIS sensor. These datasets, focused on steering angle prediction, have $12$ and $51\,\text{hr}$ of driving data, respectively. MVSEC~\cite{mvsec} presents a multimodal stereo dataset using $346\times260\,\text{px}$ DAVIS sensors along with LiDARs, IMUs, and RGB cameras for three-dimensional perception tasks, such as feature tracking, visual odometry, and stereo depth estimation. DSEC~\cite{dsec} further expands these fusion efforts by including 390K annotations for detection tasks on an hour of multimodal stereo data using $640\times480\,\text{px}$ Prophesee Gen3.1 sensors.

Prophesee~\cite{prophesee} introduced two substantial ego-motion datasets~\cite{gen1, 1megapixel} for detection. Gen1 Automotive Detection Dataset~\cite{gen1} encompasses a total of 255,781 manually annotated bounding boxes (228,123 cars and 27,658 pedestrians instances) acquired over a span of $39\,\text{hr}$ using $304\times240\,\text{px}$ Prophesee Gen1 sensor. 1~Megapixel Automotive Dataset~\cite{1megapixel} stands out as the most comprehensive ego-motion event-based detection dataset. It encompasses $15\,\text{hr}$ of recorded footage, featuring a resolution of $1280\times720\,\text{px}$, with 25M generated bounding boxes. However, they are unable to provide nighttime annotations due to their automated labeling protocol. PEDRo dataset~\cite{pedro} focuses on people detection from a robotics perspective containing 43,259 bounding boxes from 119 recordings with an average duration of $18\,\text{s}$. \\ \\
\textbf{Static perception event-based datasets} such as DVS-Pedestrian~\cite{dvspedestrian} is limited to pedestrian detection. It has 12 sequences recorded using a $346\times260\,\text{px}$ DAVIS346 camera and contains 4670 labeled instances of pedestrians. DVS-OUTLAB~\cite{dvsoutlab} explores the plausibility of using event cameras for long-time monitoring. It consists of recordings from three static $768\times640\,\text{px}$ CeleX-4 DVS event cameras featuring outdoor urban public areas involving persons, dogs, bicycles, and sports balls as objects of interest. 

\tableautorefname~\ref{tab:related_datasets} provides an overview of existing event-based datasets. Interest in event cameras has been piqued for intelligent transportation, with many datasets focused on the ego-vehicle perspective, but there is a lack of datasets focusing on traffic monitoring from a static perspective.

\begin{figure}[]
  \centering
   \includegraphics[width=0.88
\linewidth]{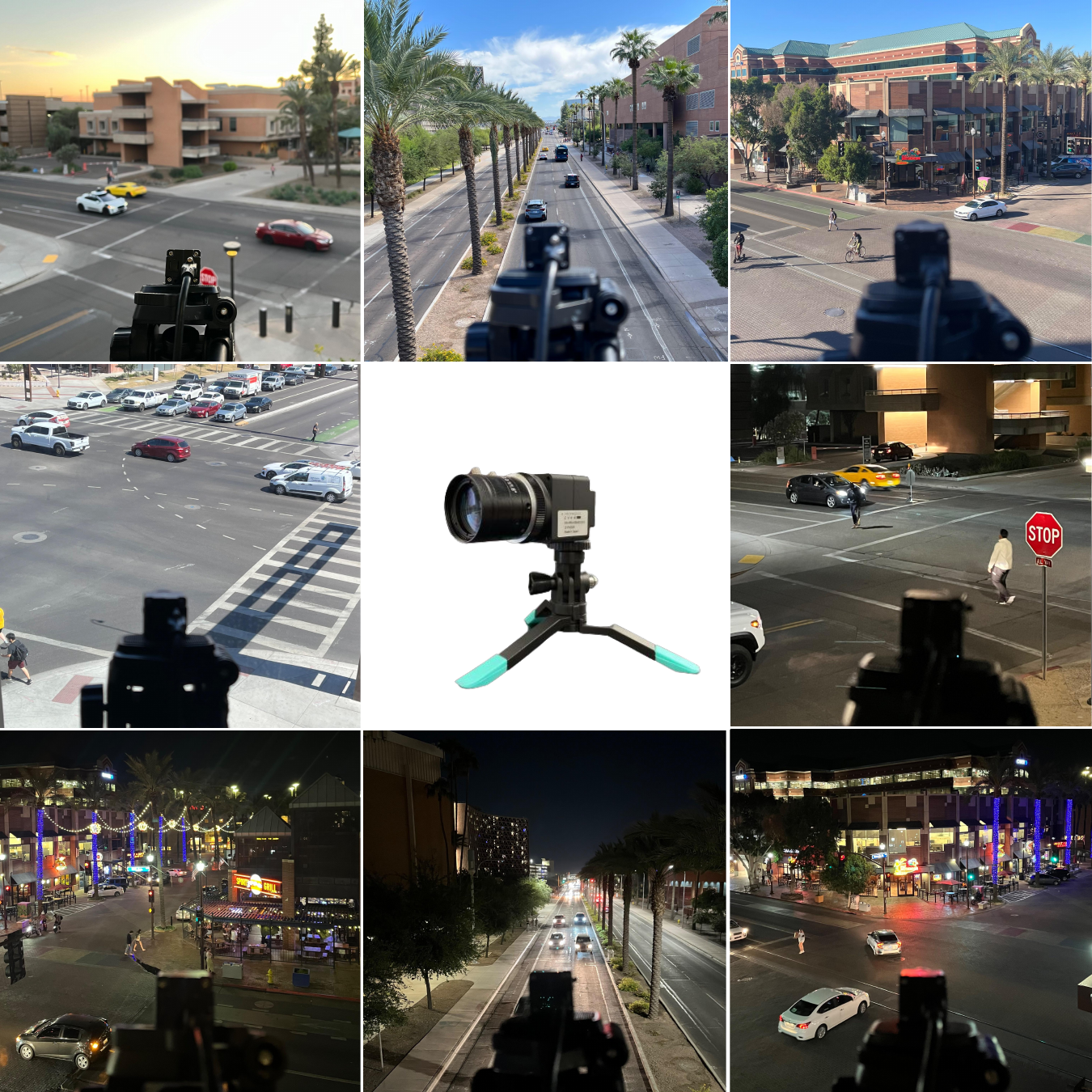}
   \caption{Data Collection Setup: The first four images from the top left display daytime data collection sites, the center image shows the Prophesee EVK4 HD event camera and the last four images depict nighttime data collection sites.}
   \label{fig02}
\end{figure}

\section{\textit{eTraM} Dataset}
\label{sec:dataset}
This section details \textit{eTraM}'s acquisition framework, preprocessing techniques, annotation, and statistics, providing insights into its applicability for traffic monitoring.

\subsection{Dataset Acquisition Framework}
We rely on the Prophesee EVK4 HD event camera~\cite{prophesee-evk4}, notable for its high resolution ($1280\times720\,\text{px}$), temporal resolution (over $10,000\,\text{fps}$), dynamic range (above $120\;\text{dB}$), and exceptional low light cutoff ($0.08\,\text{Lux}$), to capture high-quality data. The event camera was strategically positioned at approximately $6\,\text{m}$ with a pitch angle of about $35\degree{}$ to the ground. This configuration is deliberately chosen to maintain consistency with the placement of traffic cameras and to ensure comprehensive coverage of interactions between diverse traffic participants. \figureautorefname~\ref{fig02} shows different sites considered for data collection.

\textit{eTraM} comprises recorded sequences from multiple intersections, roadways, and local streets around Arizona State University, Tempe campus. The sequences were recorded for intervals of $15-30\,\text{min}$ at different times of the day, covering daytime, nighttime, and twilight. The dataset also observes different weather conditions, including sunny, overcast, and rainy. To achieve this, extensive data collection efforts were carried out over a span of $8\,\text{months}$.

\begin{figure*}[t]
\centering
\includegraphics[width=0.87\textwidth]{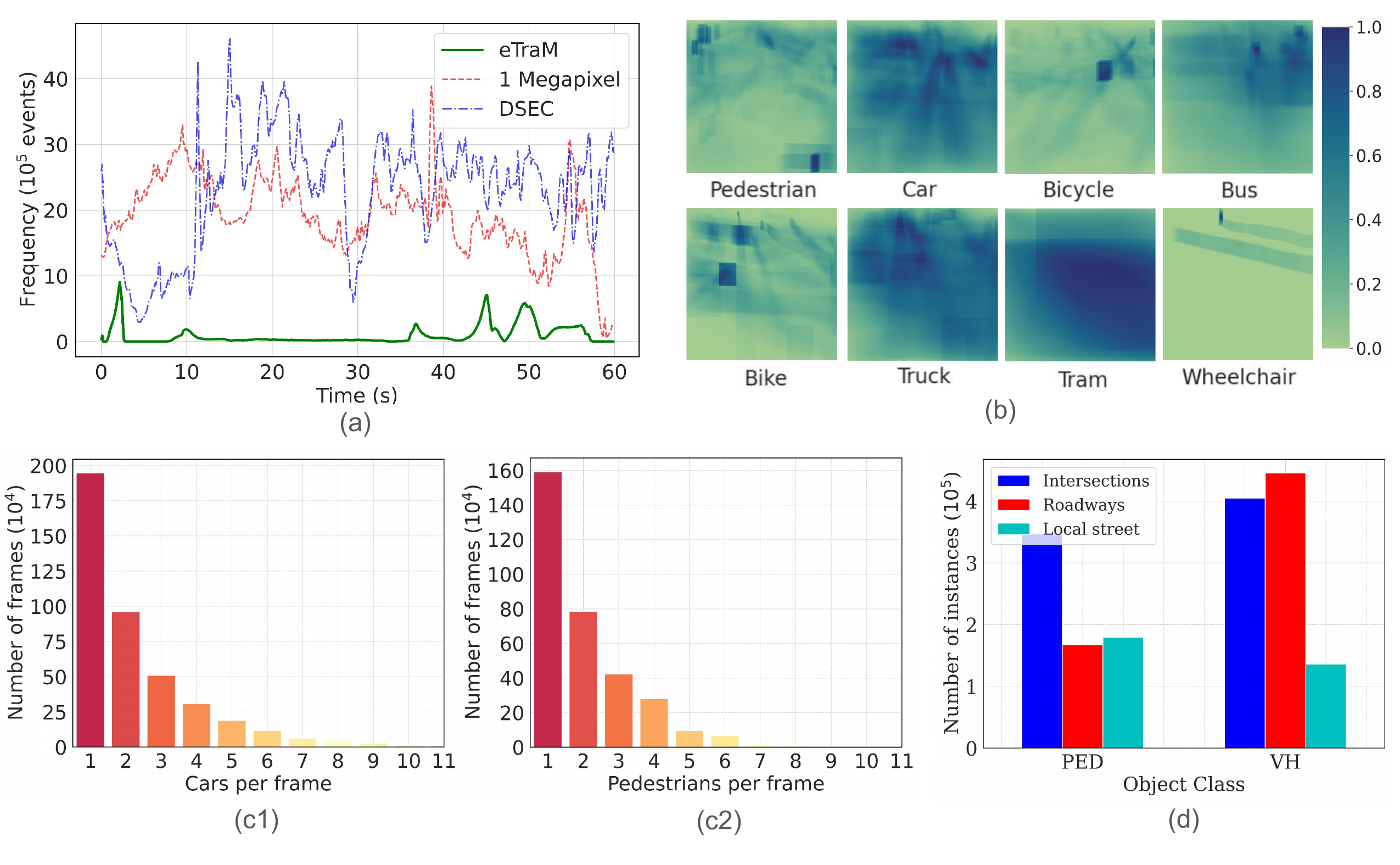}
\caption{Event-time Distribution and Object Occurrence Statistics of \textit{eTraM}: (a) Histogram of event frequency of \textit{eTraM} (static event dataset) as compared to 1 Megapixel and DSEC (ego-motion event datasets), (b) Shows the object density of various classes across the frame, (c) Power-law distribution of the number of instances within an image for most predominant classes - cars (c1) and pedestrians (c2), and (d) Distribution of two major categories across various traffic sites.}
\label{fig03}
\end{figure*}

\subsection{Preprocessing and Annotation}
Given the sensitive nature of the event sensor, it was observed that nighttime data tends to exhibit higher levels of noise, primarily attributed to reflections and pointed sources of light from streets and vehicles. To address this challenge and enhance the quality of our data, the recorded sequences are passed through a spatiotemporal filter \cite{brosch2015event}. This spatiotemporal filter works on the idea that events from real objects should occur closer together in both space and time more often compared to events from random noises~\cite{eventvisionsurvey}. 

The filtering mechanism discards an event if a threshold amount of events with the same polarity do not occur within a fixed temporal window in the vicinity of its 8-neighborhood spatial coordinates. For any $e=\langle x, y, p, t \rangle$, this condition is represented mathematically in \equationautorefname~\ref{eq:processing},
\begin{equation}
\label{eq:processing}
\sum_{t_{i} = t}^{t+\partial t} \sum_{x_{i}=x-1}^{x+1} \sum_{y_{i}=y-1}^{y+1} P (e_{i} = \langle x_{i}, y_{i}, p_{i}, t_{i}\rangle, e ) > n_{thres} 
\end{equation}
where $\partial$t represents the temporal window and P$(e_{i}, e)$ is 1 only when the polarity of $e_{i}$ and $e$ are the identical.
We choose a temporal window of $10\,\text{ms}$ with a minimum threshold of 2 neighboring events for our experiments. The specific filter values were determined through experiments detailed in \cite{dvsoutlab} and further validated through a smaller experiment conducted during the training phase. We illustrate the significance of spatiotemporal filtering through visualization in the supplementary material.

Following the denoising stage, events within the stream are partitioned into discrete time bins and consolidated into a single frame, thereby converting the asynchronous event stream into synchronous frames of $30\,\text{Hz}$. These frames are then annotated using CVAT \cite{cvat}, an open-source annotation tool. Our rigorous manual annotation process resulted in the precise identification of over 2M 2D bounding boxes.

\begin{table*}[th!]
\centering
\resizebox{0.97\textwidth}{!}{%
\normalfont
\begin{tabular}{l|c|cccc|cccc|cccc}
\toprule \midrule 
\multicolumn{1}{c|}{\multirow{2}{*}{\textbf{Traffic Site}}} &
  \multirow{2}{*}{\textbf{Lighting}} &
  \multicolumn{4}{c|}{\textbf{RVT} } &
  \multicolumn{4}{c|}{\textbf{RED}} &
  \multicolumn{4}{c}{\textbf{YOLOv8}} \\ \cline{3-14} 
\multicolumn{1}{c|}{} &
   &
  \multicolumn{1}{c}{\textbf{PED}} &
  \multicolumn{1}{c}{\textbf{VH}} &
  \multicolumn{1}{c}{\textbf{MM}} &
  \multicolumn{1}
  {c|}{\textbf{All}} &
  \multicolumn{1}{c}{\textbf{PED}} &
  \multicolumn{1}{c}{\textbf{VH}} &
  \multicolumn{1}{c}{\textbf{MM}} &
  \multicolumn{1}{c|}{\textbf{All}} &
  \multicolumn{1}
  {c}{\textbf{PED}} &
  \multicolumn{1}{c}{\textbf{VH}} &
  \multicolumn{1}{c}{\textbf{MM}} &
  \multicolumn{1}{c}{\textbf{All}}\\ \midrule \midrule 
\textbf{Intersections} &
  \multirow{4}{*}{\begin{tabular}[c]{@{}c@{}}Daytime \end{tabular}} &
  0.460 &
  0.813 &
  0.315&
  \cellcolor{highlight}\textbf{0.722}&
  0.395&
  0.593&
  0.284&
  \cellcolor{highlight}0.545&
  0.167&
  0.293 &
  0.111 &
  \cellcolor{highlight}0.190\\
\textbf{Roadways} &
   &
   0.430&
   0.733&
   0.070&
   \cellcolor{highlight}\textbf{0.627}&
  0.347 &
  0.590 &
  0.055 &
   \cellcolor{highlight}0.551&
   0.173&
   0.290&
   0.004&
   \cellcolor{highlight}0.156\\
\textbf{Local Streets} &
   &
   0.196&
   0.938&
   0.586&
   \cellcolor{highlight}0.316&
   0.208&
   0.875&
   0.695&
   \cellcolor{highlight}\textbf{0.351}&
   0.124&
   0.559&
   0.204&
   \cellcolor{highlight}0.296\\ \cmidrule{1-1} \cmidrule{3-14}
\textbf{All Scenes} &
   &
   \cellcolor{highlight2}\textbf{0.304}&
   \cellcolor{highlight2}\textbf{0.781}&
   \cellcolor{highlight2}\textbf{0.403}&
   \cellcolor{highlight3}\textbf{0.572}&
   \cellcolor{highlight2}0.302&
   \cellcolor{highlight2}0.656&
   \cellcolor{highlight2}0.251&
   \cellcolor{highlight3}0.497&
   \cellcolor{highlight2}0.142&
   \cellcolor{highlight2}0.309&
   \cellcolor{highlight2}0.112&
   \cellcolor{highlight3}0.188\\ \midrule \midrule
\textbf{Intersections} &
  \multirow{4}{*}{\begin{tabular}[c]{@{}c@{}}Nighttime \end{tabular}} &
   0.161&
   0.465&
   - &
   \cellcolor{highlight}\textbf{0.262}&
  0.149&
  0.425&
  - &
  \cellcolor{highlight}0.242&
   0.071&
   0.375&
   -&
   \cellcolor{highlight}0.149\\
\textbf{Roadways} &
   &
  0.310&
  0.827&
    -&
  \cellcolor{highlight}\textbf{0.739}&
  0.362&
  0.782&
  - &
  \cellcolor{highlight}0.726&
  0.004&
   0.229&
   -&
   \cellcolor{highlight}0.117\\
\textbf{Local Streets} &
   &
   0.739&
   0.868&
   0.097&
   \cellcolor{highlight}\textbf{0.829}&
   0.722&
   0.831&
   0.145&
  \cellcolor{highlight} 0.817&
   0.198&
   0.486&
   0.030&
   \cellcolor{highlight}0.239
   \\ \cmidrule{1-1} \cmidrule{3-14}
\textbf{All Scenes} &
    &
   \cellcolor{highlight2}\textbf{0.317}&
   \cellcolor{highlight2}\textbf{0.674}&
   \cellcolor{highlight2}0.064&
   \cellcolor{highlight3}\textbf{0.523}&
   \cellcolor{highlight2}0.303&
   \cellcolor{highlight2}0.660&
   \cellcolor{highlight2}\textbf{0.083}&
   \cellcolor{highlight3}0.504&
   \cellcolor{highlight2}0.123&
   \cellcolor{highlight2}0.322&
   \cellcolor{highlight2}0.013&
   \cellcolor{highlight3}0.153\\ \midrule \midrule 
\textbf{Overall} &
   &
   \cellcolor{highlight2}\textbf{0.309}&
   \cellcolor{highlight2}\textbf{0.717}&
   \cellcolor{highlight2}\textbf{0.313}&
   \cellcolor{highlight3}\textbf{0.539}&
   \cellcolor{highlight2}0.303&
   \cellcolor{highlight2}0.649&
   \cellcolor{highlight2}0.197&
   \cellcolor{highlight3}0.491&
   \cellcolor{highlight2}0.134&
   \cellcolor{highlight2}0.314&
   \cellcolor{highlight2}0.086&
   \cellcolor{highlight3}0.178\\ \midrule \bottomrule
\end{tabular} 
}
\caption{Baseline Evaluation: Comprehensive evaluation of state-of-the-art tensor-based methods RVT, RED, and frame-based method YOLOv8 across various traffic sites (Intersections, Roadways, Local Streets) during both daytime and nighttime for PED - Pedestrian, VH - Vehicle, and MM - Micro-mobility.}
\label{tab:baselines}
\end{table*}

\subsection{Dataset Statistics}
Here, we highlight key characteristics of the collected data and annotations. The dataset encompasses three distinct traffic monitoring scenes with $5\,\text{hr}$ of intersection, $3\,\text{hr}$ of roadway, and $2\,\text{hr}$ of local street data sequences. Data for each scene is collected at multiple locations. For instance, the intersection scene contains data from 2 four-way, three-way, and an uncontrolled intersection. Each location has daytime, nighttime, and twilight data totaling up to $10\,\text{hr}$ of data with $5\,\text{hr}$ of daytime and nighttime data.

When comparing the event distribution in \textit{eTraM} with other ego-motion event-based traffic datasets like 1~Megapixel Automotive and DSEC for a duration of $60\,\text{s}$, we observe that the number of events in \textit{eTraM} is significantly lesser by a factor of 30, as shown in \figureautorefname~\ref{fig03}~(a). This is accredited to the event camera being static in \textit{eTraM}, primarily focusing on the dynamic traffic participants in a scene. In contrast, datasets from an ego-motion perspective capture more data due to the surrounding infrastructure's relative motion and lead to dense streams of events. The sparsity of events in \textit{eTraM}, combined with the asynchronous nature of events, leads to low memory utilization, which is particularly advantageous for traffic monitoring infrastructure.

\textit{eTraM} contains over 2M instances of 2D bounding box annotations for traffic participant detection. These annotations additionally include object IDs, making it possible to evaluate multi-object tracking, as shown in the supplementary material. The annotation classes encompass a range of traffic participants, from various vehicles (cars, trucks, buses, and trams) to pedestrians and micro-mobility (bikes, bicycles, and wheelchairs). 

\figureautorefname~\ref{fig03}~(b) illustrates the spatial distribution of each class within the frame. We observe a uniform distribution of vehicle instances across the entire frame, while instances belonging to the pedestrian class cover more than 50\% of the frame. This safeguards the model from developing a bias for classes in a specific region of the frame.

For accessibility and ease of use, \textit{eTraM} is provided in multiple formats: RAW, DAT, and H5~\cite{prophesee_formats}. Additionally, annotations are available in numpy format. The dataset is split into 70\% training, 15\% validation, and 15\% testing, ensuring each subset has proportional data from each scene. Further statistics of \textit{eTraM} can be seen in \figureautorefname~\ref{fig03}~(c) \& \ref{fig03}~(d). To the best of our knowledge, this stands as a first-of-its-kind event-based dataset for traffic monitoring. Additionally, the inclusion of nighttime data enhances its versatility for a broader range of research applications.

\begin{figure*}[t!]
\centering
\includegraphics[width=0.92\textwidth]{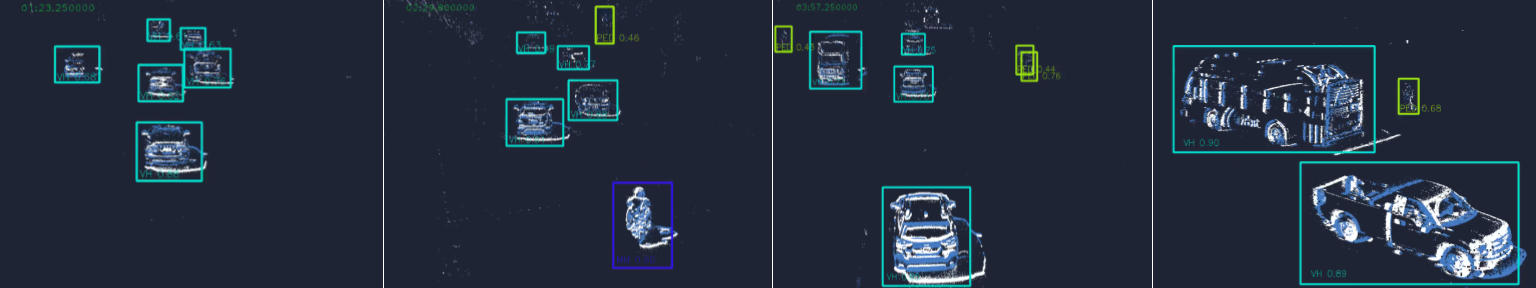}
\caption{Qualitative Results: Showcasing detection of vehicles (cyan), pedestrians (yellow), and micro-mobility (blue) on \textit{eTraM}.}
\label{fig04}
\end{figure*}

\section{Dataset Evaluation}
\label{sec:eval}
In this section, we present the results of experiments evaluating \textit{eTraM} for real-world detection scenarios. We first train and evaluate models to establish baselines and assess their performance in various scenarios (Section~\ref{sec:baseline}). We further conduct experiments to test the ability of these models to generalize on nighttime conditions and unseen scenes (Section~\ref{sec:generalization}). For all experiments, we evaluate \textit{eTraM} on SOTA tensor-based object detection methods, specifically Recurrent Vision Transformers (RVT), Recurrent Event-camera Detector (RED), and a frame-based method, You Only Look Once (YOLOv8). This comparison helps us understand how SOTA utilizing dense tensor representation, performs compared to the conventional frame-based method. For evaluation, we focus on the three broad categories of traffic participants, vehicles (VH), pedestrians (PED), and micro-mobility (MM), and report the mean Average Precision at an intersection-over-union (IoU) threshold of 50\% (AP50).

\subsection{Baseline Evaluation}
\label{sec:baseline}
To assess the performance of detection methods on \textit{eTraM}, we train them on $7\,\text{hr}$ of data and evaluate them on $1.5\,\text{hr}$ of validation and $1.5\,\text{hr}$ of test data. RVT and RED were trained from scratch over $3\,\text{days}$ on A100, while YOLOv8 was trained for $2\,\text{days}$. We conducted separate evaluations to provide insights into how each model performs in different scenes and lighting conditions. This allows us to understand how these models handle diverse and changing environmental contexts.

Several key observations emerged from the evaluation results of tensor-based models shown in \tableautorefname~\ref{tab:baselines}. Notably, the performance of vehicle detection consistently outperforms that of pedestrians across all scenes and models. The performance in the micro-mobility category exhibits significant variance, likely due to the smaller number of instances and the broad range of subjects captured within this category. During the daytime, both vehicle and pedestrian detection results are at par in intersections and roadways. In local streets, the performance of vehicle detection improves compared to other scenes due to fewer instances and reduced occlusion. Conversely, we observe a decline in the performance of pedestrian detection due to higher pedestrian densities leading to more inter-class occlusions.

During nighttime, the performance of vehicle detection in local streets and roadways remains consistent. Interestingly, pedestrian detection results improve significantly on local streets at night. This can be attributed to factors such as reduced pedestrian-to-pedestrian occlusion, instances being closer to the camera, and additional visual features like shadows that become more prominent at night due to street lights. Due to noise from various light sources, a drop in performance is observed at intersections during nighttime. Despite this, the performance during nighttime is at par with daytime scenarios, with an increase in performance on vehicle detection on roadways at night. This increase could be due to reduced inter-class occlusion as we observe fewer vehicles in nighttime conditions. 

Furthermore, tensor-based methods such as RED and RVT, which incorporate temporal information through Recurrent Neural Networks in their architecture, consistently outperform the conventional frame-based method of YOLOv8. \figureautorefname~\ref{fig04} shows an example of detection on \textit{eTraM}. The baseline results highlight the challenges and strengths of various traffic monitoring scenarios and categories, particularly showcasing the effectiveness of event-based models in nighttime conditions.

\subsection{Generalization Evaluation}
\label{sec:generalization}
A key characteristic required for real-world deployment is that models demonstrate transferability to unseen scenarios. Since event cameras are invariant to absolute illuminance levels, we expect event-based models to also demonstrate transferability to nighttime data. We conduct experiments where we control the train and test set to evaluate their transferability in the following sections quantitatively.

\subsubsection{Generalization on Night time}
In \cite{1megapixel}, qualitative assessments were conducted to compare the generalization capabilities of event-based models against frame-based models in nighttime scenarios. We aim to quantitatively assess how well event-based models RVT and RED, trained on daytime data, can perform in nighttime conditions. We conducted a controlled experiment where we trained two models for each architecture, one on $2\,\text{hr}$ of only daytime data and another with $2\,\text{hr}$ of daytime along with $45\,\text{min}$ of nighttime data. Both models were evaluated on previously unseen nighttime data.

\begin{table}[!hbt]
\centering
\resizebox{0.78\columnwidth}{!}{
\normalfont
\begin{tabular}{l|cc|cc}
\toprule \midrule
\multicolumn{1}{c|}{\multirow{2}{*}{\textbf{\begin{tabular}[c]{@{}c@{}}Train\\ Set\end{tabular}}}} &
  \multicolumn{2}{c|}{\textbf{RVT}} &
  \multicolumn{2}{c}{\textbf{RED}} \\ \cline{2-5} 
\multicolumn{1}{c|}{} &
  \multicolumn{1}{c|}{\textbf{VH}} &
  \textbf{PED} &
  \multicolumn{1}{c|}{\textbf{VH}} &
  \textbf{PED} \\ \midrule \midrule
\textbf{Day}       & \multicolumn{1}{c|}{0.566} & 0.166 & \multicolumn{1}{c|}{0.374} & 0.354 \\
\textbf{Day+Night} & \multicolumn{1}{c|}{0.761} & 0.254 & \multicolumn{1}{c|}{0.673} & 0.422 \\ \midrule \bottomrule
\end{tabular} }
\caption{Evaluation of generalization capabilities of RVT and RED on night time data for PED - pedestrian and VH - vehicle class for models trained on only daytime and a combination of day and nighttime data.}
\label{tab:day2night}
\end{table}

\begin{figure*}[!ht]
\centering
\includegraphics[width=0.92\textwidth]{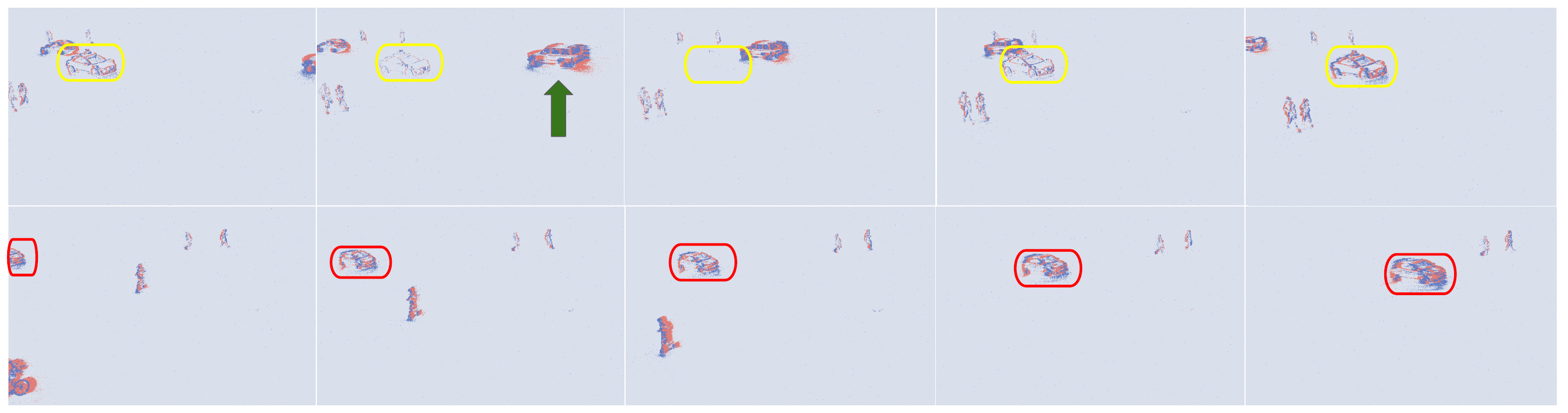}
\caption{Demonstrating Effectiveness of Event Camera for Traffic Scenarios: Yellow circle (top row) tracks a car that halts at a stop sign with a lack of motion captured in the third frame, red circle (bottom row) tracks a car that violates the stop sign where motion is continuously captured in every frame. Additionally, a green arrow (top row) shows a car traveling at a high speed, resulting in a high event density.}
\label{fig05}
\end{figure*}

The summarized results can be found in \tableautorefname~\ref{tab:day2night}. We observe a consistent trend across every object class and architecture where models trained on data with nighttime sequences consistently outperform models trained solely on daytime sequences. Despite the expectation that event cameras would exhibit proficient performance in nighttime scenarios, these observations reveal that relying solely on daytime event data may not be sufficient. These models fail to attain performance levels comparable to those achieved by models trained on nighttime data with the model. This suggests that the unique challenges posed by nighttime conditions necessitate explicit training with relevant data to ensure optimal model performance. Upon closer inspection, we find that nighttime data has environmental interferences and distinct variations of noise. This could be inherent in nighttime data due to the heightened sensitivity of event cameras, which is a factor in this discrepancy.

While event-based models are unable to generalize on nighttime data out of the box, they have at par results on daytime and nighttime data when trained on the combined dataset (refer~\tableautorefname~\ref{tab:baselines}). This highlights the need for labeled nighttime data to allow event-based models to augment current frame-based systems.

\subsubsection{Generalization on Unseen Scenes}
To substantiate the ability of event-based detectors to generalize to previously unencountered traffic scenes, we train our model on a subset of the dataset. We evaluate the ability by testing each architecture on two independent test sets, one ``held in"  that contains sequences from intersections that our model has seen during the training phase and the other on the ``held out" test set with data from an intersection that is skipped during training.

\begin{table}[!t]
\centering
\resizebox{0.78\columnwidth}{!}{
\normalfont
\begin{tabular}{l|cc|cc}
\toprule \midrule
\multicolumn{1}{c|}{\multirow{2}{*}{\textbf{\begin{tabular}[c]{@{}c@{}}Test \\ Set\end{tabular}}}} & 
  \multicolumn{2}{c|}{\textbf{RVT}} &
  \multicolumn{2}{c}{\textbf{RED}} \\ \cline{2-5} 
\multicolumn{1}{c|}{} & 
  \multicolumn{1}{c|}{\textbf{VH}} &
  \multicolumn{1}{c|}{\textbf{PED}} &
  \multicolumn{1}{c|}{\textbf{VH}} &
  \multicolumn{1}{c}{\textbf{PED}} \\ \midrule \midrule
\textbf{Held In}  & \multicolumn{1}{l|}{0.449} & 0.316 & \multicolumn{1}{l|}{0.556} & 0.521 \\ 
\textbf{Held Out} & \multicolumn{1}{l|}{0.628} & 0.529 & \multicolumn{1}{l|}{0.572} & 0.509 \\ \midrule \bottomrule
\end{tabular}  }
\caption{Evaluation of generalization capabilities of RED and RVT on unseen traffic scenarios for PED - pedestrian and VH - vehicle tested on held in and held out test set.}
\label{tab:unseen}
\end{table}

As depicted in \tableautorefname~\ref{tab:unseen}, we observe a comparable performance of the models across both major categories. On evaluating the model on the ``held out" test set, representing an entirely new and unseen traffic scene, the model's generalization capability is evident. These values are at par with the performance on the ``held in" test set. This similarity in results highlights the model's ability to seamlessly extend its learning to previously unseen traffic scenes, validating its transferability across changing environments. This generalization to unseen intersections is pivotal for real-world deployment. 

\section{Discussion}
\label{sec:discussion}

\subsection{Event Cameras in Traffic Monitoring}
In this section, we discuss the advantages of using event cameras to augment traditional frame camera systems for traffic monitoring.

Traditional cameras take continuous snapshots at a fixed frequency, potentially capturing individuals' identities and sensitive information. In comparison, event cameras register events with no color and texture information, significantly reducing the probability of gathering sensitive information of any individual~\cite{1megapixel, dvsoutlab, chakravarthi2023event}.
Further, the advantages of event cameras extend to their performance in low light conditions and their robustness towards mitigating motion blur. As we demonstrate through our baseline evaluations in \tableautorefname~\ref{tab:baselines}, event cameras display equally superior performance in nighttime conditions while maintaining a high temporal resolution that aids it in substantially reducing motion blur~\cite{8946715}.
An essential requirement of traffic monitoring is the need for resource-efficient sensors. Since event cameras are designed to operate on a sparse data stream generated by significant visual changes, they exhibit lower memory requirements and power consumption compared to traditional frame cameras~\cite{low_power, CAVIAR}. This makes event cameras sustainable and cost-effective for continuous monitoring over extended periods.

\begin{figure}[!t]
\centering
\includegraphics[width=0.95\columnwidth]{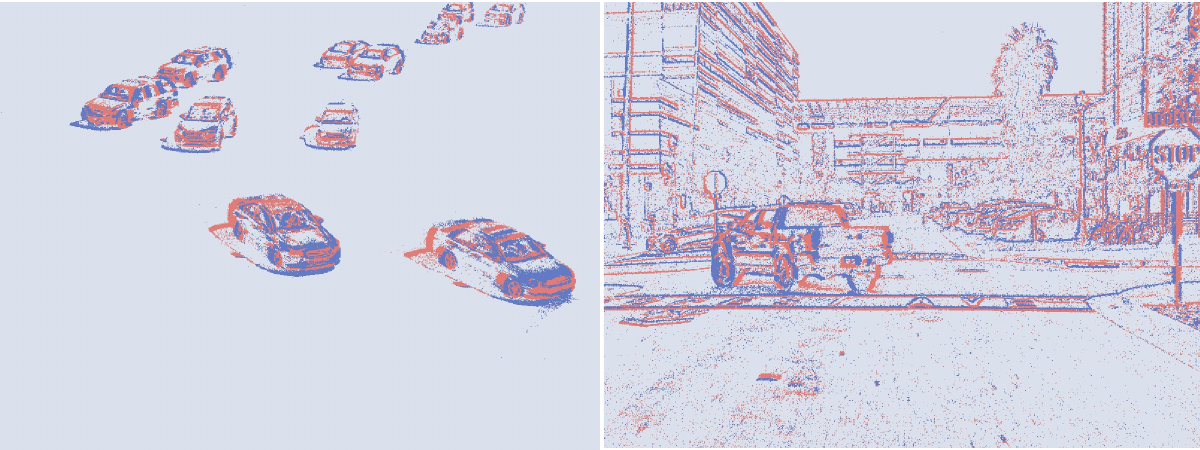}
\caption{Qualitative Comparison: Events captured from a static event camera (left) show enhanced visibility of moving vehicles compared to an ego-motion event camera (right).}
\label{fig06}
\end{figure}

\begin{figure*}[h!]
\centering
\includegraphics[width=0.8\linewidth]{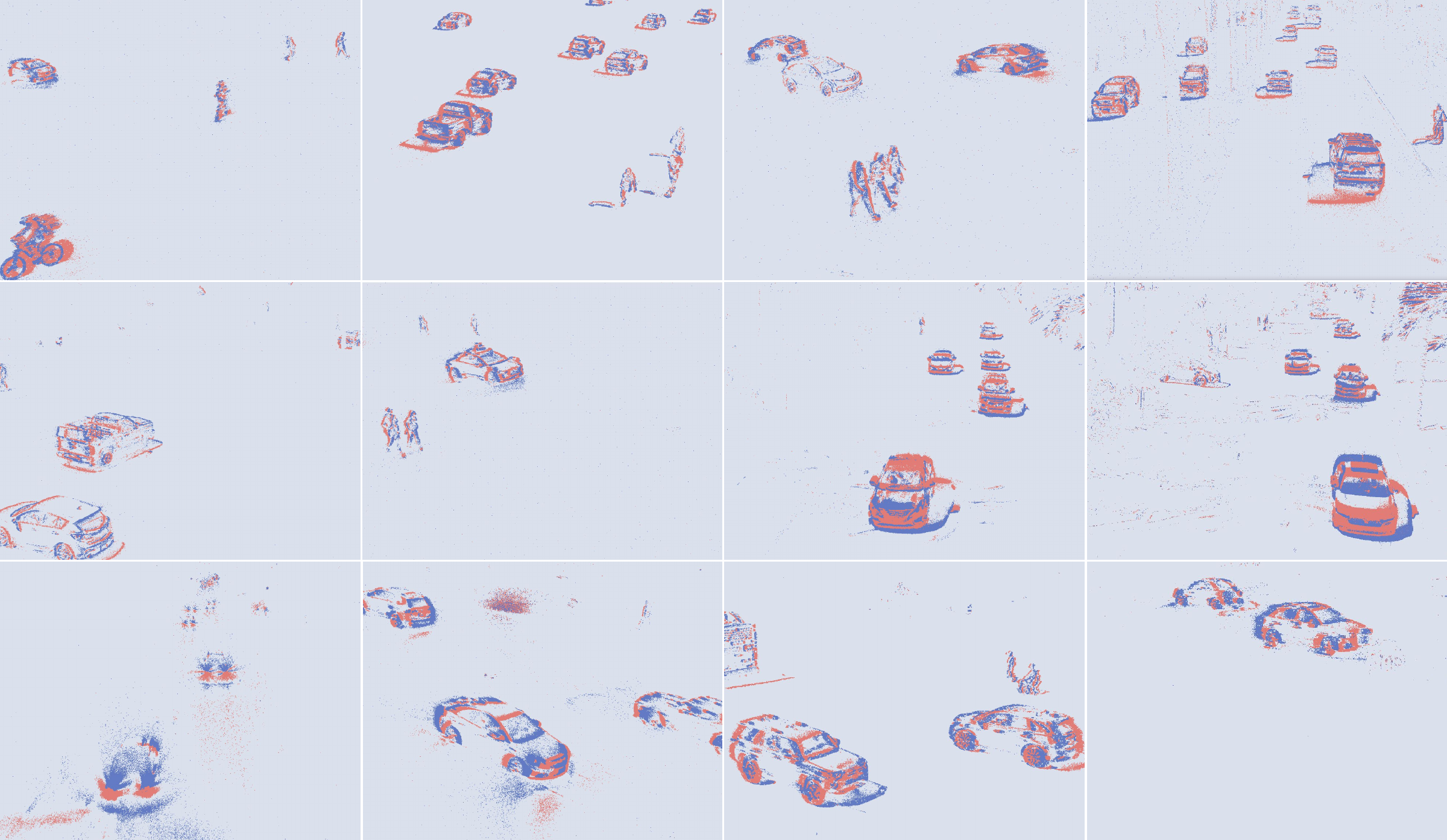}
\caption{Traffic Site Diversity in \textit{eTraM}: We show the various instances encapsulating the interactions amongst multiple traffic participants captured from a static roadside perspective with daytime (first row), twilight (second row), and nighttime (last row) showing increasing sensor noise (top to bottom) due to light sources such as headlights and street lights.}
\label{fig07}
\end{figure*}

We further highlight the capabilities brought by event cameras in traffic monitoring scenarios through two practical applications, as illustrated in \figureautorefname~\ref{fig05}. Firstly, in detecting stop sign violations, event cameras excel in capturing instantaneous changes in the visual scene, enabling precise detection of vehicles coming to a halt. The number of events associated with the object makes it easy to discern whether a moving vehicle is slowing down or has stopped. Secondly, due to the high temporal resolution of the event camera and the event density corresponding to an object, it is straightforward to detect sudden acceleration or erratic behavior of fast-moving traffic participants. This capability is critical for identifying potentially hazardous situations on the road, such as near misses, and allowing for prompt intervention or alert generation.

\subsection{Static and Ego Event-based Datasets}
Next, we discuss the difference between static and ego-motion event-based datasets, emphasizing the significance of the former for traffic monitoring. In static datasets events are more likely to be associated with relevant traffic participants rather than extraneous background changes seen in ego-motion data, enhancing the visibility of these participants. This enhanced visibility is illustrated in \figureautorefname~\ref{fig06}, which provides a qualitative comparison between events captured from a static and an ego-motion event camera. As shown, dynamic traffic participants from the static camera are better visible than those from the ego-motion camera. Another distinction is the field of view provided by static roadside datasets. Positioned at an elevated height, these cameras capture a broader scene and include far-away objects, while ego-motion datasets are constrained to the vehicle's immediate surroundings. The elevated perspective of static datasets proves advantageous, enabling the detection of traffic participants at a distance and providing a comprehensive view of the traffic environment, as seen in \figureautorefname~\ref{fig07}. This would be beneficial to a broad spectrum of applications, like traffic scene estimation~\cite{lu2023carom, wei2021learning, lei2022modeling}, road safety monitoring~\cite{srinivasan2022infrastructure} and traffic signal control~\cite{da2023llm}.

\section{Conclusion}
\label{sec:conclusion}
We present \textit{eTraM}, a large-scale manually annotated event-based dataset for traffic monitoring. Our meticulously curated dataset, captured using the cutting-edge Prophesee EVK4 HD high-resolution event camera, provides new traffic detection and tracking opportunities from a static perspective. With over $10\,\text{hr}$ of annotated event data spanning various scenes and lighting conditions, our dataset offers insights into the complex dynamics of traffic scenarios. The comprehensive annotations include over 2M bounding boxes of various traffic participants, from vehicles to pedestrians and micro-mobility. Further, nighttime generalization results highlight the value of labeled nighttime data, enabling event-based models to achieve daytime-equivalent performance on nighttime data. We demonstrate the ability of event-based models to generalize effectively to unseen scenarios, emphasizing its potential in real-world traffic monitoring. As the field of event-based sensing continues to advance, \textit{eTraM} holds the potential to serve as an invaluable resource in driving its development and presenting new opportunities in intelligent transportation systems.

\section{Acknowledgment}
This research is sponsored by NSF Pathways to Enable
Open-Source Ecosystems grant (\#2303748), the
Partnerships for Innovation grant (\#2329780), the CPS grant (\#2038666), CRII grant (\# 2153311). We thank data collection support from the Institute of Automated Mobility. 

%-------------------------------------------------------------------------

%%%%%%%%% REFERENCES
{\small
\bibliographystyle{ieee_fullname}
\bibliography{PaperForReview}

\begin{thebibliography}{10}\itemsep=-1pt

\bibitem{prophesee-evk4}
{Event Camera Evaluation Kit 4 HD IMX636 Prophesee-Sony}.

\bibitem{low_power}
Arnon Amir, Brian Taba, David Berg, Timothy Melano, Jeffrey McKinstry, Carmelo Di~Nolfo, Tapan Nayak, Alexander Andreopoulos, Guillaume Garreau, Marcela Mendoza, Jeff Kusnitz, Michael Debole, Steve Esser, Tobi Delbruck, Myron Flickner, and Dharmendra Modha.
\newblock A low power, fully event-based gesture recognition system.
\newblock In {\em 2017 IEEE Conference on Computer Vision and Pattern Recognition (CVPR)}, pages 7388--7397, 2017.

\bibitem{iam}
Arizona~Commerce Authority.
\newblock {IAM - Arizona Institute of Automated Mobility}.

\bibitem{ddd17}
Jonathan Binas, Daniel Neil, Shih-Chii Liu, and Tobi Delbruck.
\newblock Ddd17: End-to-end davis driving dataset, 2017.

\bibitem{dvsoutlab}
Tobias Bolten, Regina Pohle-Fröhlich, and Klaus~D. Tönnies.
\newblock Dvs-outlab: A neuromorphic event-based long time monitoring dataset for real-world outdoor scenarios.
\newblock In {\em 2021 IEEE/CVF Conference on Computer Vision and Pattern Recognition Workshops (CVPRW)}, pages 1348--1357, 2021.

\bibitem{pedro}
Chiara Boretti, Philippe Bich, Fabio Pareschi, Luciano Prono, Riccardo Rovatti, and Gianluca Setti.
\newblock Pedro: An event-based dataset for person detection in robotics.
\newblock In {\em Proceedings of the IEEE/CVF Conference on Computer Vision and Pattern Recognition}, pages 4064--4069, 2023.

\bibitem{brosch2015event}
Tobias Brosch, Stephan Tschechne, and Heiko Neumann.
\newblock On event-based optical flow detection.
\newblock {\em Frontiers in neuroscience}, 9:137, 2015.

\bibitem{nuscenes}
Holger Caesar, Varun Bankiti, Alex~H. Lang, Sourabh Vora, Venice~Erin Liong, Qiang Xu, Anush Krishnan, Yu Pan, Giancarlo Baldan, and Oscar Beijbom.
\newblock nuscenes: A multimodal dataset for autonomous driving.
\newblock In {\em Proceedings of the IEEE/CVF Conference on Computer Vision and Pattern Recognition (CVPR)}, June 2020.

\bibitem{chakravarthi2023event}
Bharatesh Chakravarthi, M Manoj~Kumar, and BN Pavan~Kumar.
\newblock Event-based sensing for improved traffic detection and tracking in intelligent transport systems toward sustainable mobility.
\newblock In {\em International Conference on Interdisciplinary Approaches in Civil Engineering for Sustainable Development}, pages 83--95. Springer, 2023.

\bibitem{AVevent}
Guang Chen, Hu Cao, Jorg Conradt, Huajin Tang, Florian Rohrbein, and Alois Knoll.
\newblock Event-based neuromorphic vision for autonomous driving: A paradigm shift for bio-inspired visual sensing and perception.
\newblock {\em IEEE Signal Processing Magazine}, 37(4):34--49, 2020.

\bibitem{da2023llm}
Longchao Da, Minchiuan Gao, Hao Mei, and Hua Wei.
\newblock Prompt to transfer: Sim-to-real transfer for traffic signal control with prompt learning.
\newblock {\em Proceedings of the Thirty-Eighth AAAI Conference on Artificial Intelligence (AAAI'24)}, 2024.

\bibitem{gen1}
Pierre de Tournemire, Davide Nitti, Etienne Perot, Davide Migliore, and Amos Sironi.
\newblock A large scale event-based detection dataset for automotive, 2020.

\bibitem{du2023safelight}
Wenlu Du, Junyi Ye, Jingyi Gu, Jing Li, Hua Wei, and Guiling Wang.
\newblock Safelight: A reinforcement learning method toward collision-free traffic signal control.
\newblock In {\em Proceedings of the AAAI conference on artificial intelligence}, volume~37, pages 14801--14810, 2023.

\bibitem{1597116}
Li Fei-Fei, R. Fergus, and P. Perona.
\newblock One-shot learning of object categories.
\newblock {\em IEEE Transactions on Pattern Analysis and Machine Intelligence}, 28(4):594--611, 2006.

\bibitem{eventvisionsurvey}
Guillermo Gallego, Tobi Delbr{\"{u}}ck, Garrick Orchard, Chiara Bartolozzi, Brian Taba, Andrea Censi, Stefan Leutenegger, Andrew~J. Davison, J{\"{o}}rg Conradt, Kostas Daniilidis, and Davide Scaramuzza.
\newblock Event-based vision: {A} survey.
\newblock {\em CoRR}, abs/1904.08405, 2019.

\bibitem{garcia2021safety}
Marichelo Garcia-Venegas, Diego~Alberto Mercado-Ravell, Luis~A Pinedo-Sanchez, and Carlos~A Carballo-Monsivais.
\newblock On the safety of vulnerable road users by cyclist detection and tracking.
\newblock {\em Machine Vision and Applications}, 32(5):109, 2021.

\bibitem{dsec}
Mathias Gehrig, Willem Aarents, Daniel Gehrig, and Davide Scaramuzza.
\newblock Dsec: A stereo event camera dataset for driving scenarios.
\newblock {\em IEEE Robotics and Automation Letters}, 2021.

\bibitem{kitti}
Andreas Geiger, Philip Lenz, and Raquel Urtasun.
\newblock Are we ready for autonomous driving? the kitti vision benchmark suite.
\newblock In {\em 2012 IEEE Conference on Computer Vision and Pattern Recognition}, pages 3354--3361, 2012.

\bibitem{ddd20}
Yuhuang Hu, Jonathan Binas, Daniel Neil, Shih-Chii Liu, and Tobi Delbruck.
\newblock Ddd20 end-to-end event camera driving dataset: Fusing frames and events with deep learning for improved steering prediction, 2020.

\bibitem{v2e}
Yuhuang Hu, Shih-Chii Liu, and Tobi Delbruck.
\newblock v2e: From video frames to realistic dvs events.
\newblock In {\em Proceedings of the IEEE/CVF Conference on Computer Vision and Pattern Recognition (CVPR) Workshops}, pages 1312--1321, June 2021.

\bibitem{726791}
Y. Lecun, L. Bottou, Y. Bengio, and P. Haffner.
\newblock Gradient-based learning applied to document recognition.
\newblock {\em Proceedings of the IEEE}, 86(11):2278--2324, 1998.

\bibitem{lei2022modeling}
Xiaoliang Lei, Hao Mei, Bin Shi, and Hua Wei.
\newblock Modeling network-level traffic flow transitions on sparse data.
\newblock In {\em Proceedings of the 28th ACM SIGKDD Conference on Knowledge Discovery and Data Mining}, pages 835--845, 2022.

\bibitem{liu2013adaptive}
Siyuan Liu, Yisong Yue, and Ramayya Krishnan.
\newblock Adaptive collective routing using gaussian process dynamic congestion models.
\newblock In {\em Proceedings of the 19th ACM SIGKDD international conference on Knowledge discovery and data mining}, pages 704--712, 2013.

\bibitem{lu2023carom}
Duo Lu, Eric Eaton, Matt Weg, Wei Wang, Steven Como, Jeffrey Wishart, Hongbin Yu, and Yezhou Yang.
\newblock Carom air-vehicle localization and traffic scene reconstruction from aerial videos.
\newblock In {\em 2023 IEEE International Conference on Robotics and Automation (ICRA)}, pages 10666--10673. IEEE, 2023.

\bibitem{mei2023libsignal}
Hao Mei, Xiaoliang Lei, Longchao Da, Bin Shi, and Hua Wei.
\newblock Libsignal: An open library for traffic signal control.
\newblock {\em Machine Learning}, pages 1--37, 2023.

\bibitem{dvspedestrian}
Shu Miao, Guang Chen, Xiangyu Ning, Yang Zi, Kejia Ren, Zhenshan Bing, and Alois Knoll.
\newblock Neuromorphic vision datasets for pedestrian detection, action recognition, and fall detection.
\newblock {\em Frontiers in Neurorobotics}, 13, 2019.

\bibitem{orchard2015converting}
Garrick Orchard, Ajinkya Jayawant, Gregory Cohen, and Nitish Thakor.
\newblock Converting static image datasets to spiking neuromorphic datasets using saccades, 2015.

\bibitem{1megapixel}
Etienne Perot, Pierre de Tournemire, Davide Nitti, Jonathan Masci, and Amos Sironi.
\newblock Learning to detect objects with a 1 megapixel event camera.
\newblock In H. Larochelle, M. Ranzato, R. Hadsell, M.F. Balcan, and H. Lin, editors, {\em Advances in Neural Information Processing Systems}, volume~33, pages 16639--16652. Curran Associates, Inc., 2020.

\bibitem{prophesee}
Prophesee.
\newblock {Prophesee - Metavision for Machines}.

\bibitem{prophesee_formats}
{Prophesee S.A}.
\newblock {Prophesee Documentation - File Formats}.
\newblock © Copyright Prophesee S.A - All Rights Reserved.

\bibitem{esim}
Henri Rebecq, Daniel Gehrig, and Davide Scaramuzza.
\newblock Esim: an open event camera simulator.
\newblock In Aude Billard, Anca Dragan, Jan Peters, and Jun Morimoto, editors, {\em Proceedings of The 2nd Conference on Robot Learning}, volume~87 of {\em Proceedings of Machine Learning Research}, pages 969--982. PMLR, 29--31 Oct 2018.

\bibitem{8946715}
Henri Rebecq, René Ranftl, Vladlen Koltun, and Davide Scaramuzza.
\newblock High speed and high dynamic range video with an event camera.
\newblock {\em IEEE Transactions on Pattern Analysis and Machine Intelligence}, 43(6):1964--1980, 2021.

\bibitem{UASMonitoring}
J.P. Rodríguez-Gomez, A.~Gómez Eguíluz, J.R. Martínez-de Dios, and A. Ollero.
\newblock Asynchronous event-based clustering and tracking for intrusion monitoring in uas.
\newblock In {\em 2020 IEEE International Conference on Robotics and Automation (ICRA)}, pages 8518--8524, 2020.

\bibitem{cvat}
Boris Sekachev, Nikita Manovich, Maxim Zhiltsov, Andrey Zhavoronkov, Dmitry Kalinin, Ben Hoff, TOsmanov, Dmitry Kruchinin, Artyom Zankevich, DmitriySidnev, Maksim Markelov, Johannes222, Mathis Chenuet, a andre, telenachos, Aleksandr Melnikov, Jijoong Kim, Liron Ilouz, Nikita Glazov, Priya4607, Rush Tehrani, Seungwon Jeong, Vladimir Skubriev, Sebastian Yonekura, vugia truong, zliang7, lizhming, and Tritin Truong.
\newblock opencv/cvat: v1.1.0, Aug. 2020.

\bibitem{CAVIAR}
Rafael Serrano-Gotarredona, Matthias Oster, Patrick Lichtsteiner, Alejandro Linares-Barranco, Rafael Paz-Vicente, Francisco Gomez-Rodriguez, Luis Camunas-Mesa, Raphael Berner, Manuel Rivas-Perez, Tobi Delbruck, Shih-Chii Liu, Rodney Douglas, Philipp Hafliger, Gabriel Jimenez-Moreno, Anton Civit~Ballcels, Teresa Serrano-Gotarredona, Antonio~J. Acosta-Jimenez, and BernabÉ Linares-Barranco.
\newblock Caviar: A 45k neuron, 5m synapse, 12g connects/s aer hardware sensory–processing– learning–actuating system for high-speed visual object recognition and tracking.
\newblock {\em IEEE Transactions on Neural Networks}, 20(9):1417--1438, 2009.

\bibitem{srinivasan2022infrastructure}
Anshuman Srinivasan, Yoga Mahartayasa, Varun~Chandra Jammula, Duo Lu, Steven Como, Jeffrey Wishart, Yezhou Yang, and Hongbin Yu.
\newblock Infrastructure-based lidar monitoring for assessing automated driving safety.
\newblock Technical report, SAE Technical Paper, 2022.

\bibitem{waymo}
Pei Sun, Henrik Kretzschmar, Xerxes Dotiwalla, Aurelien Chouard, Vijaysai Patnaik, Paul Tsui, James Guo, Yin Zhou, Yuning Chai, Benjamin Caine, Vijay Vasudevan, Wei Han, Jiquan Ngiam, Hang Zhao, Aleksei Timofeev, Scott Ettinger, Maxim Krivokon, Amy Gao, Aditya Joshi, Yu Zhang, Jonathon Shlens, Zhifeng Chen, and Dragomir Anguelov.
\newblock Scalability in perception for autonomous driving: Waymo open dataset.
\newblock {\em CoRR}, abs/1912.04838, 2019.

\bibitem{9812059}
Abhishek Tomy, Anshul Paigwar, Khushdeep~S. Mann, Alessandro Renzaglia, and Christian Laugier.
\newblock Fusing event-based and rgb camera for robust object detection in adverse conditions.
\newblock In {\em 2022 International Conference on Robotics and Automation (ICRA)}, pages 933--939, 2022.

\bibitem{wei2021learning}
Hua Wei, Chacha Chen, Chang Liu, Guanjie Zheng, and Zhenhui Li.
\newblock Learning to simulate on sparse trajectory data.
\newblock In {\em Machine Learning and Knowledge Discovery in Databases: Applied Data Science Track: European Conference, ECML PKDD 2020, Ghent, Belgium, September 14--18, 2020, Proceedings, Part IV}, pages 530--545. Springer, 2021.

\bibitem{wei2021we}
Hua Wei, Dongkuan Xu, Junjie Liang, and Zhenhui~Jessie Li.
\newblock How do we move: Modeling human movement with system dynamics.
\newblock In {\em Proceedings of the AAAI Conference on Artificial Intelligence}, volume~35, pages 4445--4452, 2021.

\bibitem{zhang2023vehicle}
Zhuang Zhang, Lijun Zhang, Dejian Meng, Luying Huang, Wei Xiao, and Wei Tian.
\newblock Vehicle kinematics-based image augmentation against motion blur for object detectors.
\newblock Technical report, SAE Technical Paper, 2023.

\bibitem{eventvisionsurvey2}
Xu Zheng, Yexin Liu, Yunfan Lu, Tongyan Hua, Tianbo Pan, Weiming Zhang, Dacheng Tao, and Lin Wang.
\newblock Deep learning for event-based vision: A comprehensive survey and benchmarks, 2023.

\bibitem{zhou2020enhancing}
Fan Zhou, Liang Li, Ting Zhong, Goce Trajcevski, Kunpeng Zhang, and Jiahao Wang.
\newblock Enhancing urban flow maps via neural odes.
\newblock In {\em Proceedings of the Twenty-Ninth International Joint Conference on Artificial Intelligence,$\{$IJCAI$\}$ 2020}, 2020.

\bibitem{mvsec}
Alex~Zihao Zhu, Dinesh Thakur, Tolga Özaslan, Bernd Pfrommer, Vijay Kumar, and Kostas Daniilidis.
\newblock The multivehicle stereo event camera dataset: An event camera dataset for 3d perception.
\newblock {\em IEEE Robotics and Automation Letters}, 3(3):2032--2039, 2018.

\end{thebibliography}


\begin{thebibliography}{1}\itemsep=-1pt

\bibitem{prophesee-evk4}
{Event Camera Evaluation Kit 4 HD IMX636 Prophesee-Sony}.

\bibitem{iou_tracking}
Erik Bochinski, Volker Eiselein, and Thomas Sikora.
\newblock High-speed tracking-by-detection without using image information.
\newblock In {\em International Workshop on Traffic and Street Surveillance for Safety and Security at IEEE AVSS 2017}, Lecce, Italy, Aug. 2017.

\bibitem{brosch2015event}
Tobias Brosch, Stephan Tschechne, and Heiko Neumann.
\newblock On event-based optical flow detection.
\newblock {\em Frontiers in neuroscience}, 9:137, 2015.

\bibitem{rvt}
Mathias Gehrig and Davide Scaramuzza.
\newblock Recurrent vision transformers for object detection with event cameras, 2023.

\bibitem{YOLOV8}
Glenn Jocher, Ayush Chaurasia, and Jing Qiu.
\newblock {YOLO by Ultralytics}, Jan. 2023.

\bibitem{timesurface1}
Xavier Lagorce, Garrick Orchard, Francesco Galluppi, Bertram~E. Shi, and Ryad~B. Benosman.
\newblock Hots: A hierarchy of event-based time-surfaces for pattern recognition.
\newblock {\em IEEE Transactions on Pattern Analysis and Machine Intelligence}, 39(7):1346--1359, 2017.

\bibitem{histo2}
Ana~I. Maqueda, Antonio Loquercio, Guillermo Gallego, Narciso Garcia, and Davide Scaramuzza.
\newblock Event-based vision meets deep learning on steering prediction for self-driving cars.
\newblock In {\em 2018 IEEE/CVF Conference on Computer Vision and Pattern Recognition}. IEEE, June 2018.

\bibitem{histo1}
Diederik~Paul Moeys, Federico Corradi, Emmett Kerr, Philip Vance, Gautham Das, Daniel Neil, Dermot Kerr, and Tobi Delbruck.
\newblock Steering a predator robot using a mixed frame/event-driven convolutional neural network, 2016.

\bibitem{1megapixel}
Etienne Perot, Pierre de Tournemire, Davide Nitti, Jonathan Masci, and Amos Sironi.
\newblock Learning to detect objects with a 1 megapixel event camera.
\newblock In H. Larochelle, M. Ranzato, R. Hadsell, M.F. Balcan, and H. Lin, editors, {\em Advances in Neural Information Processing Systems}, volume~33, pages 16639--16652. Curran Associates, Inc., 2020.

\end{thebibliography}
}

\end{document}

\title{\textit{eTraM}: Event-based Traffic Monitoring Dataset \\
Supplementary Material}

\author{Aayush Atul Verma$^*$, Bharatesh Chakravarthi$^*$, Arpitsinh Vaghela$^*$, Hua Wei, Yezhou Yang \\
Arizona State University \\
{\tt\small \{averma90, bshettah, avaghel3, hua.wei, yz.yang\}@asu.edu}\
}

\twocolumn[{%
\renewcommand\twocolumn[1][]{#1}%
\maketitle
}]

\def\thefootnote{*}\footnotetext{Equal contribution}

\section{\textit{eTraM} Statistics}
\label{sec:stats}
This section provides additional statistics about our dataset for a more comprehensive understanding of \textit{eTraM}. \textit{eTraM} consists of 10 hours of data collected from the Prophesee EVK4 HD camera~\cite{prophesee-evk4}. Beyond the annotated static perception data, \textit{eTraM} includes sequences of ego-motion event-based data, offering increased dataset diversity and experimentation opportunities. 

\begin{figure}[!hbt]
\centering
\includegraphics[width=1\columnwidth]{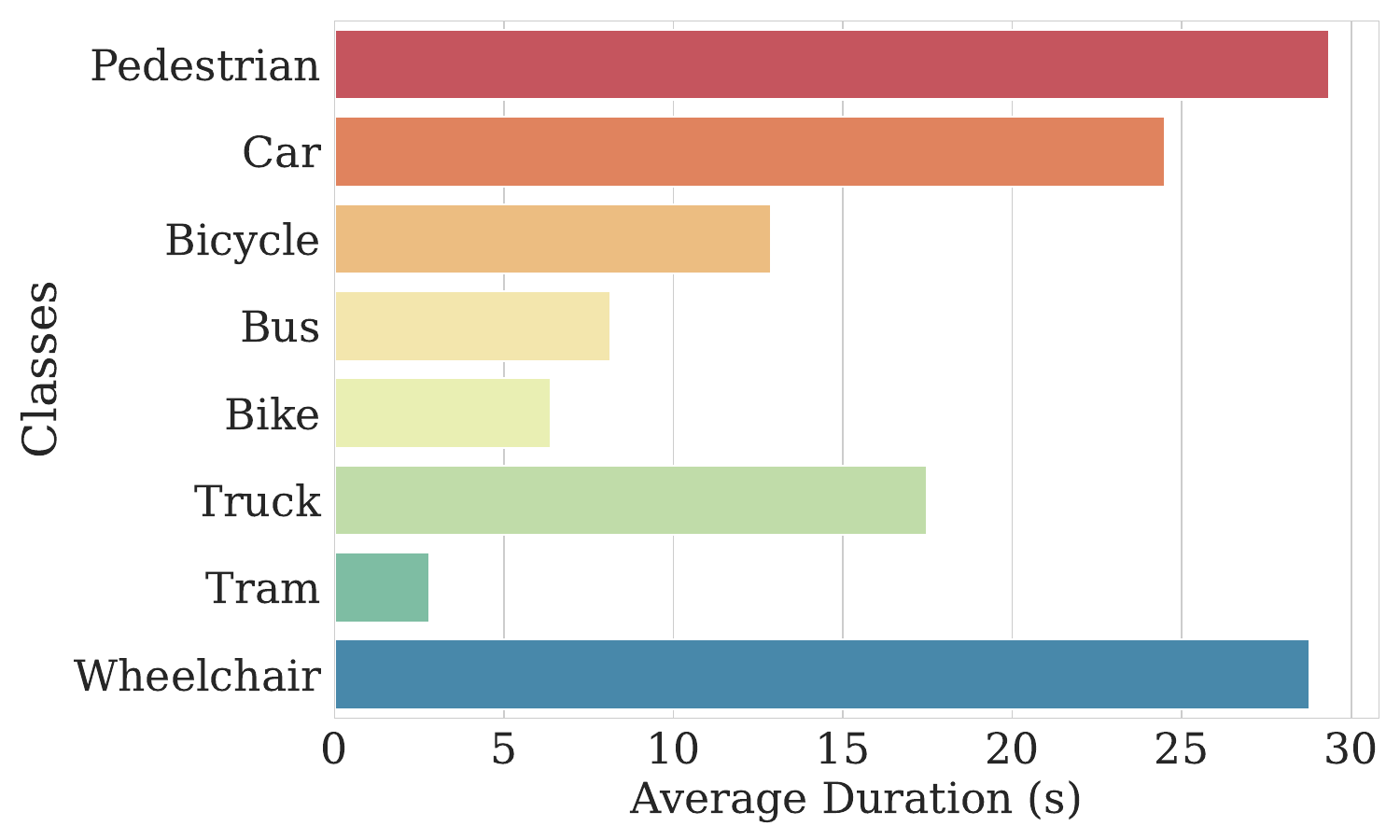}
\caption{Average Duration Spent by Objects from Each Class: The bar plot illustrates the average duration, in seconds, spent by objects of different classes, providing insights into the temporal characteristics of each class in the dataset.}
\label{fig:duration_by_class}
\end{figure}

Figure~\ref{fig:duration_by_class} presents the average duration spent by instances from each class at the traffic sites. This temporal analysis sheds light on the distinctive time dynamics of different classes within the dataset. Participants from the pedestrian and wheelchair classes spend the maximum time at the traffic sites, correlating with their respective movement speeds. In contrast, classes within the vehicle category tend to spend relatively less time in comparison.

\begin{figure}[!hbt]
    \centering
\includegraphics[width=0.85\columnwidth]{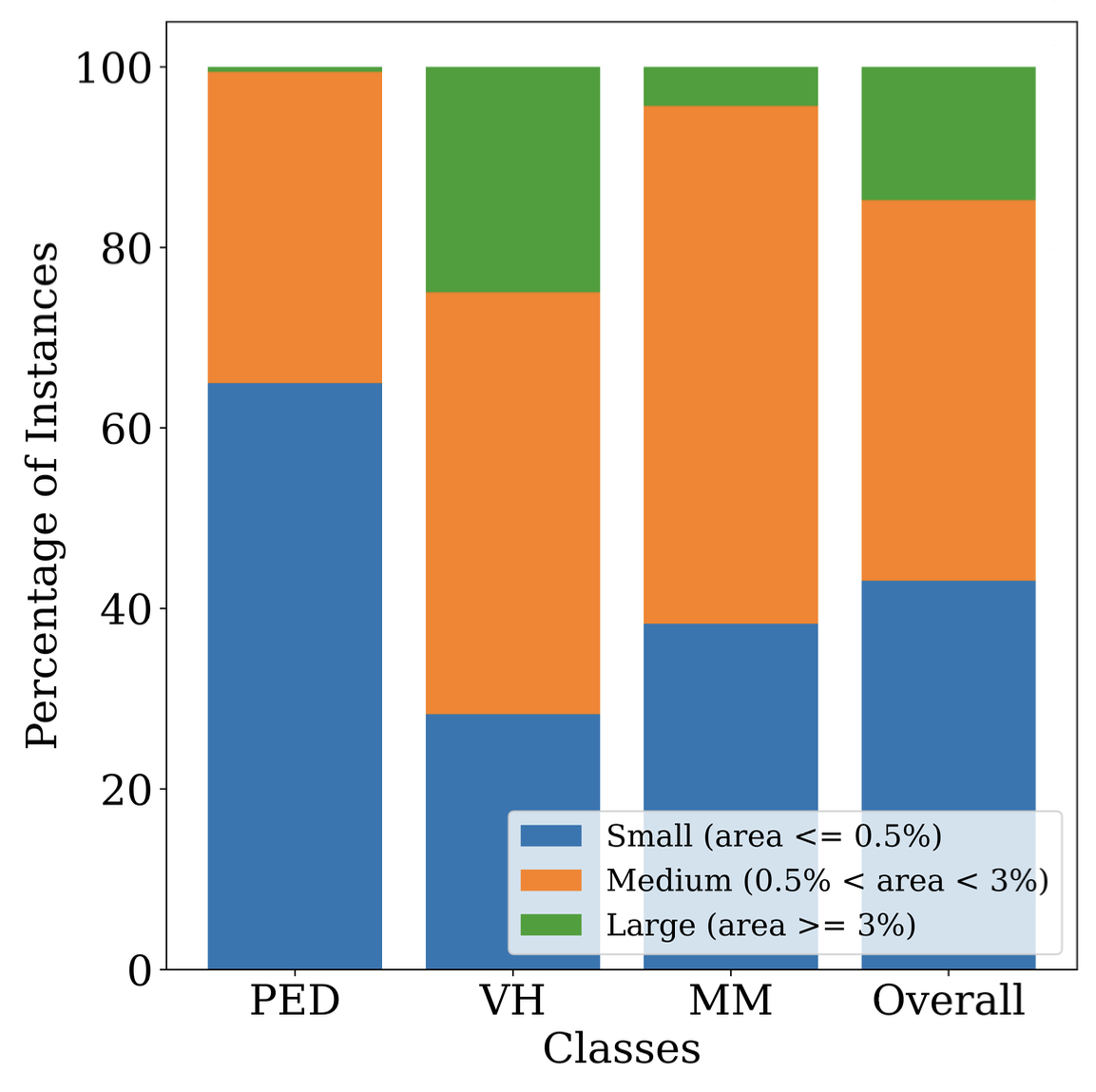}
    \caption{Analysis of the distribution of objects categorized by size (small, medium, and large)}
\label{fig:size_wise_grouping}
\end{figure}

Further, we analyze the distribution of different categories (VH, PED,
and MM) by the area they cover - small, medium, and large, as shown in \figureautorefname~\ref{fig:size_wise_grouping}.

\begin{table}[!hbt]
\centering
\resizebox{0.95\columnwidth}{!}{%
\begin{tabular}{l|cccc|cccc}
\toprule \midrule
\multicolumn{1}{c|}{\multirow{2}{*}{\textbf{\begin{tabular}[c]{@{}c@{}}Object \\ Size\end{tabular}}}} &
  \multicolumn{4}{c|}{\textbf{RVT}} &
  \multicolumn{4}{c}{\textbf{RED}} \\ \cline{2-9} 
\multicolumn{1}{c|}{} &
  \multicolumn{1}{c|}{\textbf{PED}} &
  \multicolumn{1}{c|}{\textbf{VH}} &
  \multicolumn{1}{c|}{\textbf{MM}} &
  \textbf{All} &
  \multicolumn{1}{c|}{\textbf{PED}} &
  \multicolumn{1}{c|}{\textbf{VH}} &
  \multicolumn{1}{c|}{\textbf{MM}} &
  \textbf{All} \\ \midrule \midrule
\textbf{Small} &
  \multicolumn{1}{c|}{0.308} &
  \multicolumn{1}{c|}{0.705} &
  \multicolumn{1}{c|}{\textbf{0.276}} & 0.516
   &
  \multicolumn{1}{c|}{0.324} &
  \multicolumn{1}{c|}{0.556} &
  \multicolumn{1}{c|}{\textbf{0.274}} & 0.385
   \\
\textbf{Medium} &
 \multicolumn{1}{c|}{\textbf{0.859}} &
  \multicolumn{1}{c|}{\textbf{0.722}} &
  \multicolumn{1}{c|}{0.100} &\textbf{0.722}
   &
  \multicolumn{1}{c|}{\textbf{0.661}} &
  \multicolumn{1}{c|}{\textbf{0.763}} &
  \multicolumn{1}{c|}{0.159} & 0.561
   \\
\textbf{Large} &
  \multicolumn{1}{c|}{-} &
  \multicolumn{1}{c|}{0.637} &
  \multicolumn{1}{c|}{-} & 0.637
   &
  \multicolumn{1}{c|}{-} &
  \multicolumn{1}{c|}{0.701} &
  \multicolumn{1}{c|}{-} & \textbf{0.701}
\\ \midrule \bottomrule  
\end{tabular}%
}
\caption{Evaluation of object size impact on the performance of RVT and RED.}
\label{tab:obj-size}
\end{table} 

\begin{figure*}[!t]
\centering
\includegraphics[width=0.95\textwidth]{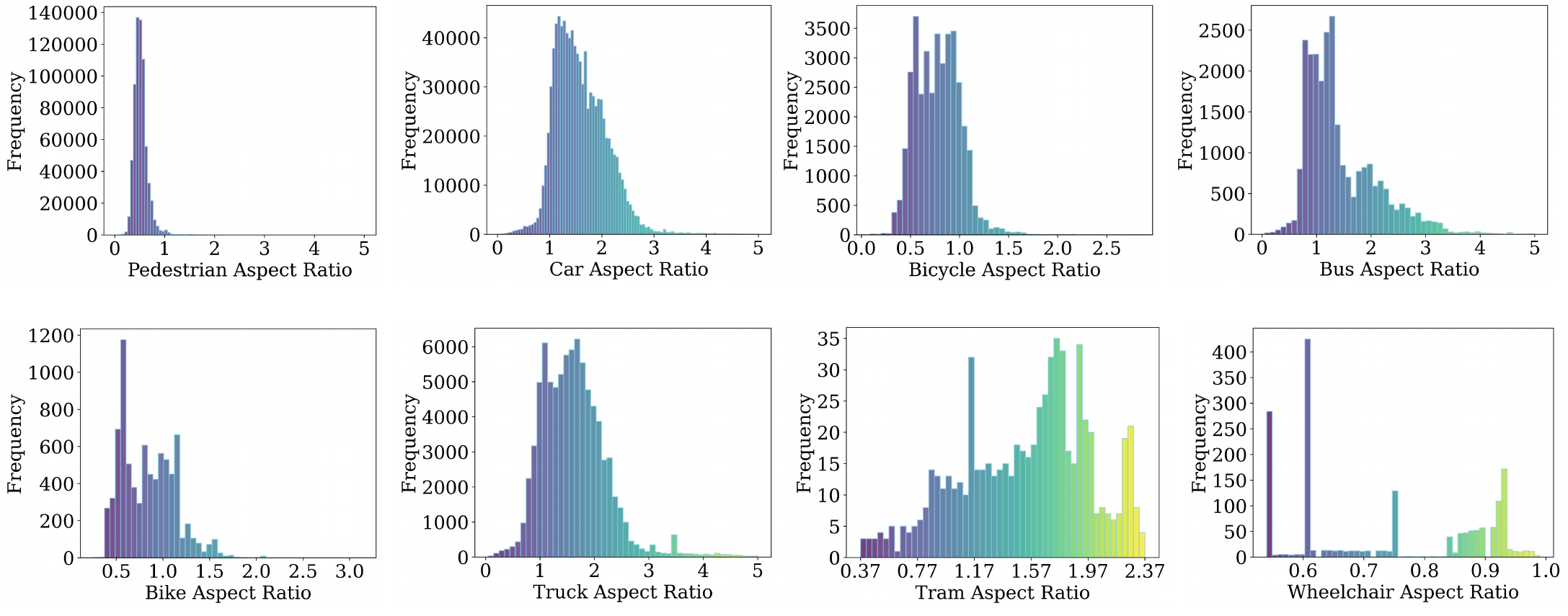}
\caption{Aspect Ratio Distribution in \textit{eTraM}: A histogram depicting the frequency distribution of aspect ratios across different classes in \textit{eTraM}, providing a comprehensive overview of the dataset's characteristics.}
\label{fig:asp_ratio}
\end{figure*}

\begin{figure*}[!th]
\centering
\includegraphics[width=0.77\textwidth]{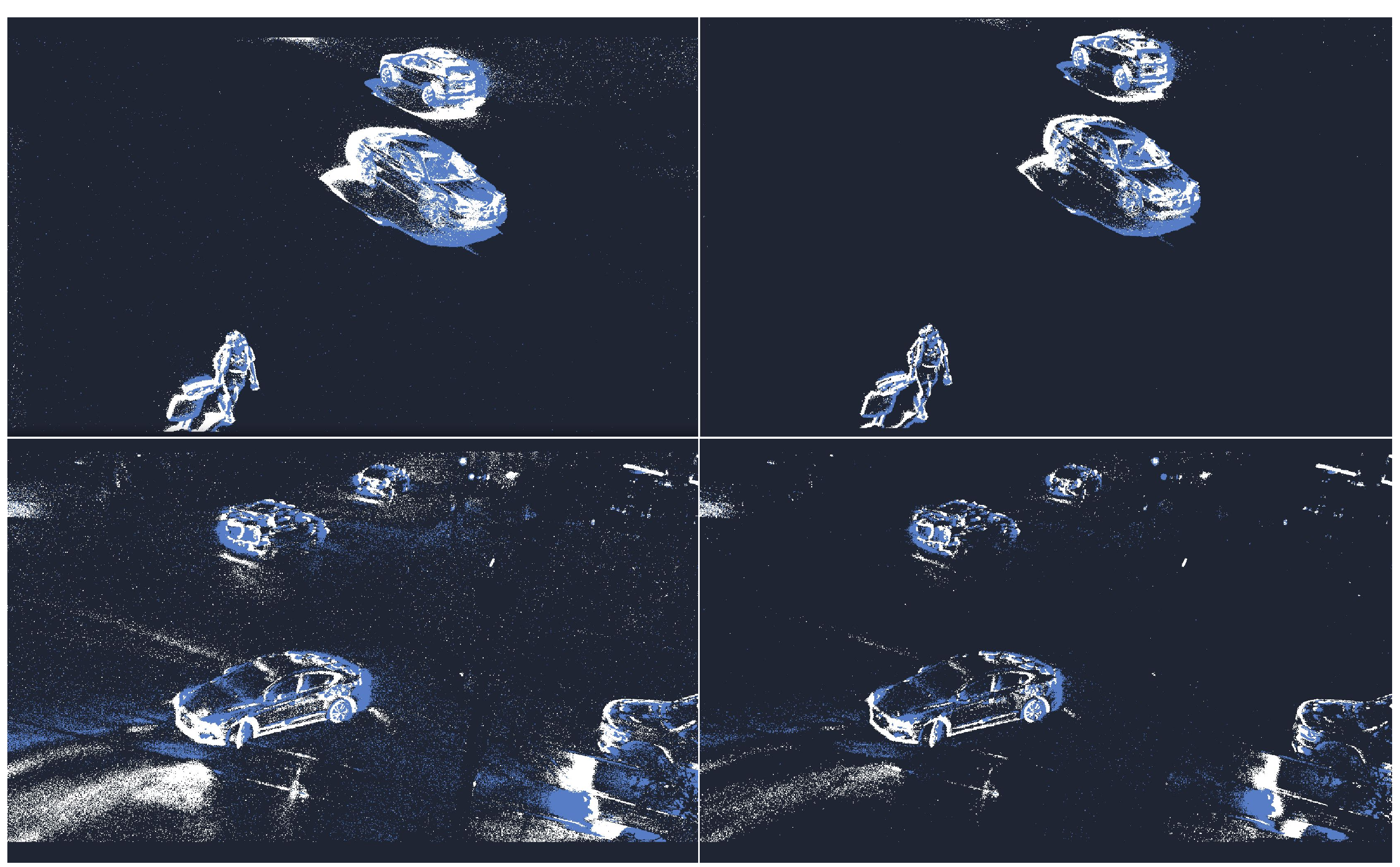}
\caption{Impact of Spatiotemporal Filtering on Event Camera Data: Comparison of a noisy pre-filtered image (left) and the enhanced clarity achieved post-filtering (right) on daytime (top row) and nighttime data (bottom row).}
\label{fig:denoising}
\end{figure*}

Based on the size classification, we also establish benchmarks in Table~\ref{tab:obj-size}. Upon analysis, it becomes evident that both models exhibit similar trends in performance. Specifically, the performance on instances categorized as medium-sized within the pedestrian and vehicle categories is consistently superior to that on small and large-sized instances of their category. Although the performance on vehicles tends to be similar performance across all three size classifications, the performance in the pedestrian category observes a significant drop when evaluated with small-sized instances. In contrast, performance on small-sized instances is better than medium-sized for micro-mobility. However, the results of micro-mobility in its best-performing size classification are still worse than the worst-performance of pedestrian and vehicle categories.
These results signify a performance degradation when dealing with small-sized objects, particularly micro-mobility. This limitation may stem from the lack of contour and color information in raw event data.

Additionally, Figure~\ref{fig:asp_ratio} shows the frequency of aspect ratios for each class in \textit{eTraM}.

\begin{figure*}[!thb]
\centering
\includegraphics[width=\textwidth]{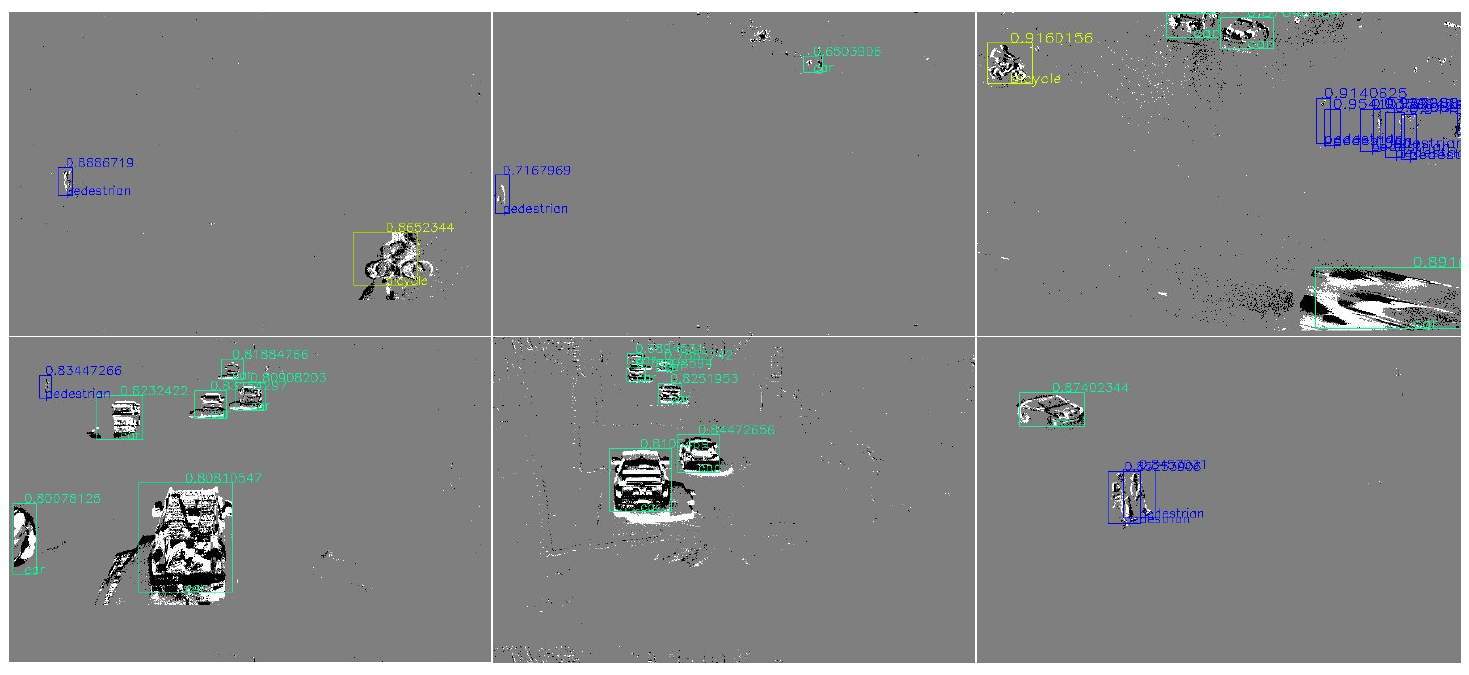}
\caption{Traffic Participant Object Detection by RVT: Snapshots illustrating the detection results of RVT at various traffic sites, showcasing its performance in diverse real-world scenarios.}
\label{fig:rvt_det} 
\includegraphics[width=\textwidth]
{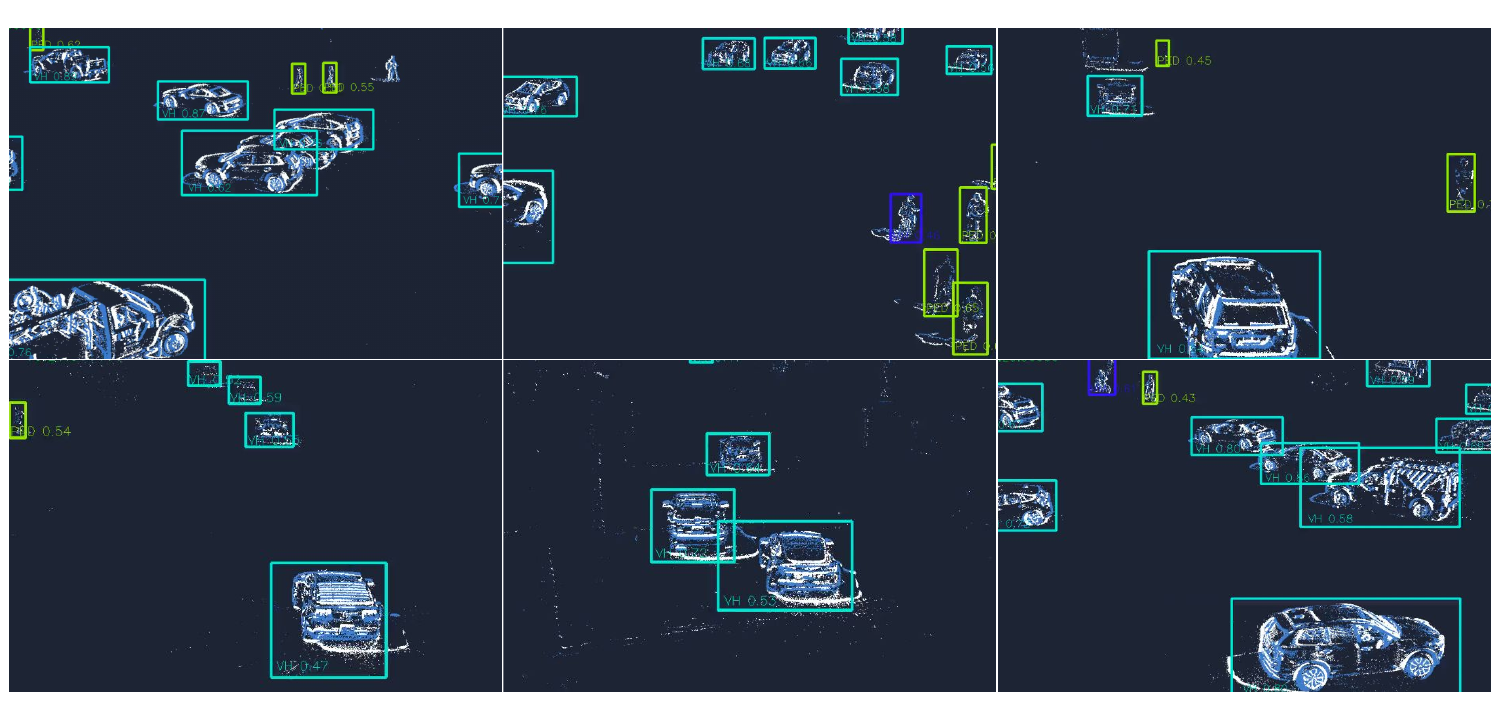}
\caption{Traffic Participant Object Detection by RED. Snapshots illustrate the detection results of RED at various traffic sites, showcasing its performance in diverse real-world scenarios.}
\label{fig:red_det} 
\end{figure*}

\begin{figure}[!t]
\centering
\includegraphics[width=\columnwidth]{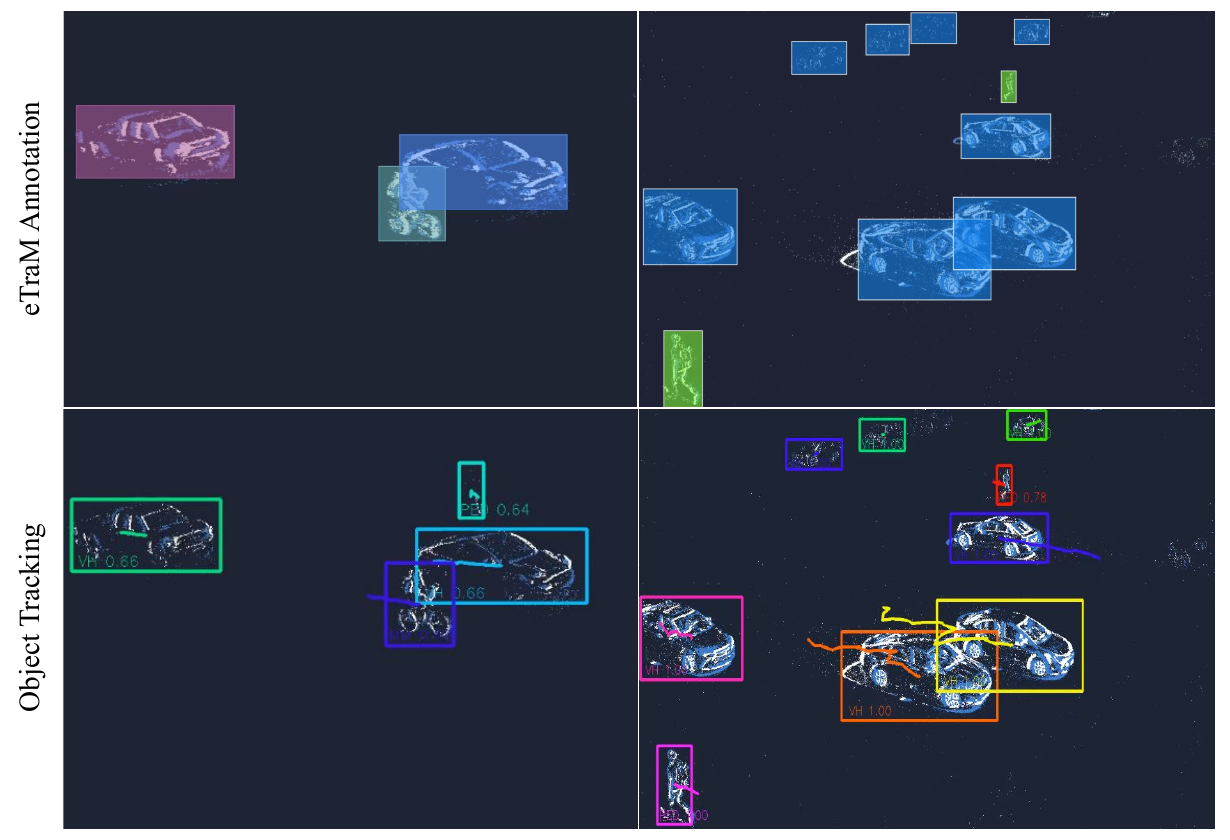}
\caption{Illustration of Intersection-over-Union based Multi-Object Tracking on the detection results of RVT}
\label{fig:tracking}
\end{figure}

\section{Denoising Using Spatiotemporal Filter}
\label{sec:spatiotemporal}
To address the noise present in the event stream, particularly heightened during nighttime data with increased levels of reflections and pointed light sources from streets and vehicles, a denoising step is implemented for \textit{eTraM}.

\figureautorefname~\ref{fig:denoising} qualitatively illustrates the effectiveness of the spatiotemporal filter~\cite{brosch2015event} by presenting a side-by-side comparison of images before and after applying the filter, showcasing the impact of noise reduction on event data frames.

\section{Implementation Details}
\label{sec:implementation_dets}
To assess how well event-based models perform on \textit{eTraM}, we trained the state-of-the-art architectures - RVT~\cite{rvt}, RED~\cite{1megapixel}, and YOLOv8~\cite{YOLOV8} on $7\,\text{hr}$ of data. We evaluated them on $1.5\,\text{hr}$ of validation and $1.5\,\text{hr}$ of test data to establish the baselines. Learning rates of $2\times10^{-6}$, $2\times10^{-4}$, and $1\times10^{-2}$ are chosen, respectively. 

\subsection{Input Preprocessing}
In this section, we define the input representations used, namely the Histogram of Events~\cite{histo1, histo2} and Time Surfaces~\cite{timesurface1}. The following representations were used to establish baselines and conduct the generalization experiments.

% \subsubsection{Histogram of Events}
\textbf{Histogram of Events} involves assigning each event to a specific cell based on its position $(x, y)$ and a time bin determined by its timestamp $(t)$. Subsequently, the total count of events is tallied within each cell and time bin, with separate counts for each polarity recorded in distinct output channels. This process results in a total of two output channels.

Let $H$ represent a four-dimensional tensor with dimensions $n, c, h, w$, where $n$ represents the index of the timestamp, $c$ represents the channel for the two polarities, $h$ represents the height, and $w$ represents the width of the input event stream. Every new event $\langle x, y, p, t\rangle$ corresponds to a specific histogram decided by the time bin that the timestamp corresponds to. Next, the histogram is updated by adding 1 at the spatial coordinates of the new event. The mathematical representation of the update is as shown in \equationautorefname~\ref{eq:histo_processing}.
This four-dimensional input representation was used by the tensor-based approaches - RVT and RED.

\begin{equation}
\label{eq:histo_processing}
 H(\tfrac{t}{\Delta}, p, y, x) =  H(\tfrac{t}{\Delta}, p, y, x) + 1 
\end{equation}

\textbf{Time Surface}, an alternative event processing method, involves recording the timestamp of the most recently received event for each pixel. This technique considers polarities independently, resulting in the output of two channels.

We incorporate an exponential decay to the timestamps to diminish the influence of older events.
Assuming $t_0 = 0$ for simplicity, this decay process is implemented.
The input representation is represented as a three-dimensional tensor $\langle p, w, h\rangle$, where $p$ represents the polarity, $h$ represents the height, and $w$ represents the width of the input event stream.

For each event $\langle x, y, p, t\rangle$ when $t \leq t_{i}$, its contribution to the time surface at time $t_{i}$ can be mathematically represented as shown in Equation~\ref{eq:ts_processing}. The two polarities were considered as the input channels for YOLOv8, and the architecture was updated accordingly.
\begin{equation}
   \label{eq:ts_processing}
 TS_{t_{i}}(p, y, x) = exp(-\tfrac{t_{i}-t}{\tau})
\end{equation}

\section{Detection and Tracking Examples}
\label{sec:detections}
This section features illustrations of detections by the tensor-based methods - RVT (Figure~\ref{fig:rvt_det}) and RED (Figure~\ref{fig:red_det}) across the various traffic scenarios within \textit{eTraM}. 

The detection results are been used to perform tracking using an IoU-based thresholding technique~\cite{iou_tracking}. This results in a Multi-Object Tracking Accuracy (MOTA)/Multi-Object Tracking Precision value (MOTP) of 0.18/0.28 on \textit{eTraM}'s test set. It is worth reiterating that the precise evaluation of tracking performance is made possible solely through the inclusion of object IDs within \textit{eTraM}. An example of ground truth objects and their corresponding tracking is illustrated in \figureautorefname~\ref{fig:tracking}.

%%%%%%%%% REFERENCES
{\small
\bibliographystyle{ieee_fullname}
\bibliography{supp}
}